\RequirePackage{fix-cm}
\documentclass[twocolumn]{svjour3}          
\smartqed  
\usepackage{bbm,dsfont}
\usepackage{amsmath,amsfonts}
\usepackage[ruled,vlined]{algorithm2e}
\usepackage{array}
\usepackage[caption=false,font=normalsize,labelfont=sf,textfont=sf]{subfig}
\usepackage{url}
\usepackage{verbatim}
\usepackage{graphicx}
\usepackage{cite}
\usepackage{arydshln}
\usepackage{multirow}
\usepackage{adjustbox}
\usepackage{bbding}
\usepackage{multirow}
\usepackage{booktabs}
\usepackage{makecell}
\usepackage[pdftex,
CJKbookmarks=true, bookmarksnumbered=true, bookmarksopen=true,
colorlinks=true,
pdfborder=001,
citecolor=blue, linkcolor=blue, anchorcolor=red, urlcolor=black
]{hyperref}
\usepackage{booktabs}
\usepackage{amssymb}

\usepackage{marvosym}

\usepackage{natbib}
\usepackage{hyphenat}

\usepackage[dvipsnames]{xcolor} 
\usepackage{pifont}
\journalname{International Journal of Computer Vision}
\renewcommand{\arraystretch}{1.3}

\definecolor{darkmag}{rgb}{0.55,0,0.55}
\begin{document}
\begin{sloppypar}

\title{Semantic-Aligned Learning with Collaborative Refinement for Unsupervised VI-ReID}



%

\titlerunning{Semantic-Aligned Learning with Collaborative Refinement for Unsupervised VI-ReID}   


\author{De Cheng$^{\ast}$ 
        \thanks{$^{\ast}$ Equal contribution.}
        \and
        Lingfeng He$^{\ast}$ \and
        Nannan Wang\textsuperscript{\Letter}
        \and
        Dingwen Zhang \and
        Xinbo Gao \and
}


\institute{
De Cheng$^{\ast}$, Equal contribution,
\at
Xidian University, Xi'an 710071, China \\
\email{\href{mailto:dcheng@xidian.edu.cn}{\textcolor{blue}{dcheng@xidian.edu.cn}}}
\and
Lingfeng He$^{\ast}$, Equal contribution,
\at
Xidian University, Xi'an 710071, China \\
\email{\href{mailto:lfhe@stu.xidian.edu.cn}{\textcolor{blue}{lfhe@stu.xidian.edu.cn}}} 
\and
Nannan Wang\textsuperscript{\Letter}, Corresponding authors \at
Xidian University, Xi'an 710071, China \\
\email{\href{mailto:nnwang@xidian.edu.cn}{\textcolor{blue}{nnwang@xidian.edu.cn}}}
\and
Dingwen Zhang \at
Northwestern Polytechnical University, Xi'an 710071, China \\
\email{\href{mailto:zdw2006yyy@nwpu.edu.cn}{\textcolor{blue}{zdw2006yyy@nwpu.edu.cn}}}
\and
Xinbo Gao \at
Chongqing University of Posts and Telecommunications, Chongqing 400065, China \\
\email{\href{mailto:gaoxb@cqupt.edu.cn}{\textcolor{blue}{gaoxb@cqupt.edu.cn}}}
}

\date{Received: date / Accepted: date}

\maketitle

\begin{abstract}

Unsupervised visible-infrared person re-identification (USL-VI-ReID) seeks to match pedestrian images of the same individual across different modalities without human annotations for model learning. Previous methods unify pseudo-labels of cross-modality images through label association algorithms and then design contrastive learning framework for global feature learning. However, these methods overlook the cross-modality variations in feature representation and pseudo-label distributions brought by fine-grained patterns. This insight results in insufficient modality-shared learning when only global features are optimized. To address this issue, we propose a Semantic-Aligned Learning with Collaborative Refinement (SALCR) framework, which builds up optimization objective for specific fine-grained patterns emphasized by each modality, thereby achieving complementary alignment between the label distributions of different modalities. Specifically, we first introduce a Dual Association with Global Learning (DAGI) module to unify the pseudo-labels of cross-modality instances in a bi-directional manner. Afterward, a Fine-Grained Semantic-Aligned Learning (FGSAL) module is carried out to explore part-level semantic-aligned patterns emphasized by each modality from cross-modality instances. Optimization objective is then formulated based on the semantic-aligned features and their corresponding label space. To alleviate the side-effects arising from noisy pseudo-labels, we propose a Global-Part Collaborative Refinement (GPCR) module to mine reliable positive sample sets for the global and part features dynamically and optimize the inter-instance relationships. Extensive experiments demonstrate the effectiveness of the proposed method, which achieves superior performances to state-of-the-art methods. Our code is available at \href{https://github.com/FranklinLingfeng/code-for-SALCR}{\textcolor{blue}{https://github.com/FranklinLingfeng/code-for-SALCR}}.

\end{abstract}

\section{Introduction}


Visible-Infrared Person Re-identification (VI-ReID) aims to match the same pedestrian captured by both visible and infrared cameras \cite{MPANet, SSFT, cm-SSFT, agw, SYSU-MM01, RegDB}, which serves as a supplement to the single-modality ReID, improving generalization under poor illumination conditions and enabling effective night-time surveillance system.
It has garnered significant attention in recent years due to its wide applications in intelligent surveillance systems. 
Although existing VI-ReID methods \cite{SAAI, IDKL, CAL, PartMix} have achieved remarkable performance, they heavily rely on extensive human-annotated training data, which is more time-consuming and expensive than manual annotations in single-modality ReID. 
Therefore, unsupervised VI-ReID (USL-VI-ReID) has emerged as an important research field,
aiming to retrieve individuals from cross-modality cameras without any annotations for training.
This technology is more accommodating to real-world applications due to its economical cost.

\begin{figure}
\centering
\includegraphics[width=0.48\textwidth]{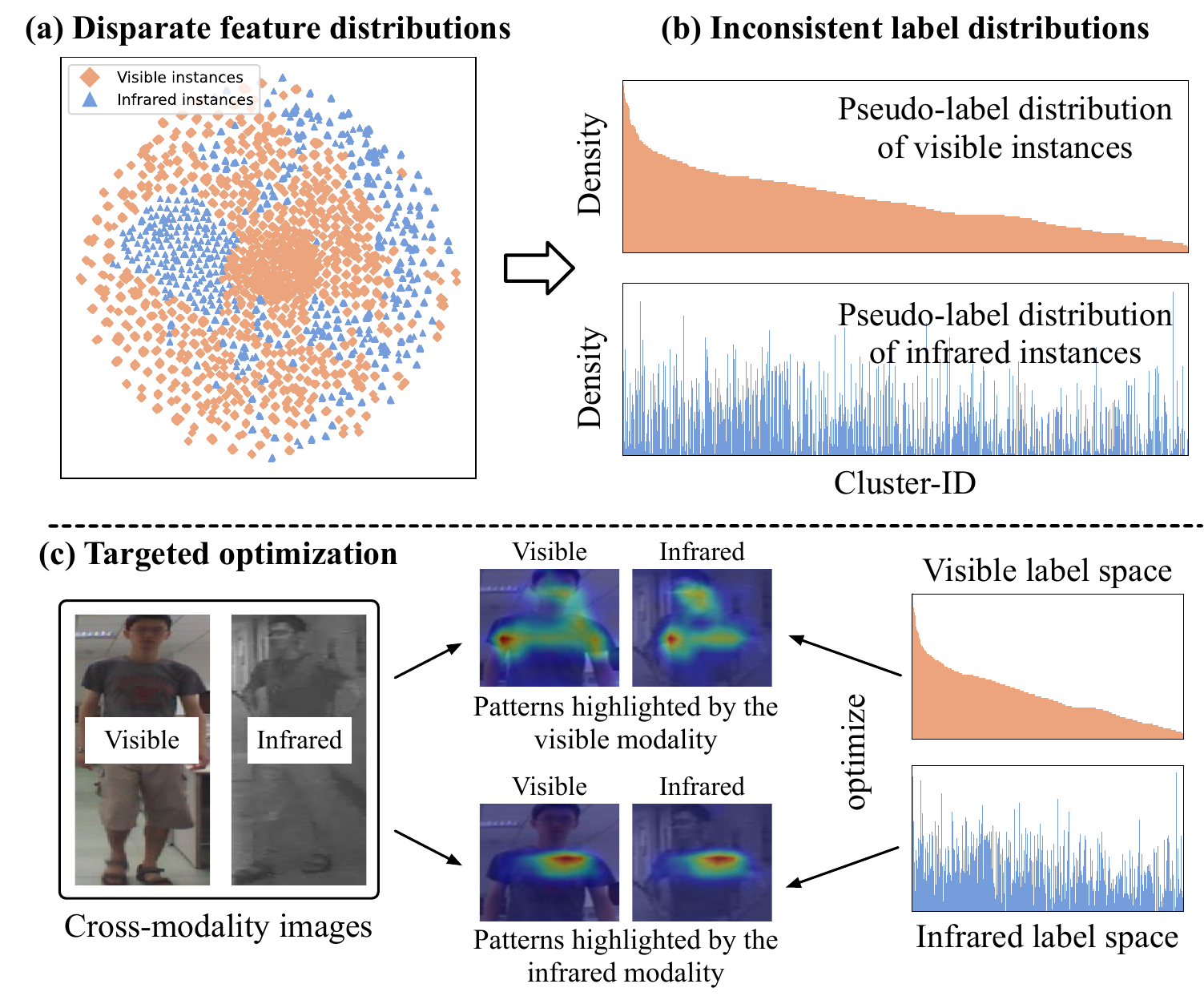}
\caption{Illustration of our idea. 
(a) illustrates the distinct feature distributions observed within different modalities, where features from different modalities lie in different subspaces.
(b) showcases the inconsistency between pseudo-label distributions of instances within different modalities.
(c) depicts our motivation: to optimize the fine-grained patterns emphasized by each modality through their respective label space.
}\label{fig:motivation}
\vspace{-4mm}
\end{figure}

For the USL-VI-ReID, the inter-modality representation discrepancies make it more challenging and distinct from unsupervised single-modality ReID.
The conventional clustering-based methods \cite{PPLR, ISE, ICE, ClusterContrast} fail to generate reliable modality-unified pseudo-labels due to persistent cross-modality clustering inconsistencies.
Several USL-VI-ReID algorithms \cite{PGM, MBCCM, DOTLA, MMM, GUR, DCCL} alleviate this issue by unifying the label space among cross-modality instances and designing contrastive learning framework to learn modality-invariant representations.
However, these methods primarily rely on global features and overlook the cross-modality variations in feature representation and pseudo-label distributions brought by fine-grained features.
The intrinsic dissimilarities between visible and infrared imagery prompt the model to attend differently to specific fine-grained patterns within each modality during intra-modality learning. 
Consequently, this results in disparate feature distributions among cross-modality instances (Fig.\ref{fig:motivation}(a)), with visible images emphasizing detailed color variations and infrared images focusing more on general features like edges and contours. 
Such distinctions in feature distributions are also reflected by the inconsistency between visible label space (pseudo-labels of visible instances) and infrared label space, as shown in Fig.\ref{fig:motivation}(b).
Fig. \ref{fig:motivation}(b) show the pseudo-label distributions of visible and infrared instances in SYSU-MM01 \cite{SYSU-MM01}.
“Density” represents the number of instances within each cluster. We observed that certain clusters consist almost exclusively of images from a single modality. 
In datasets with relatively balanced class distributions (e.g., SYSU-MM01), this phenomenon can be attributed to certain clusters capturing fine-grained features that are prominent in one modality but less distinguishable in the other.
This observation motivates us to explore the specific semantic-aligned patterns of cross-modality images emphasized by each modality, and then optimize them through pseudo-labels from their respective specific label space, as shown in Fig.\ref{fig:motivation}(c).
Such optimization objective paradigm highlights distinct fine-grained patterns in different label spaces, thus achieving the complementary between inconsistent label distributions within different modalities.

In pursuit of this objective, we propose Semantic-Aligned Learning with Collaborative Refinement (SALCR), a novel framework designed to explore and learn fine-grained semantic-aligned patterns salient in each modality.
Initially, we set up a Dual Association with Global Learning (DAGL) module to assign pseudo-labels in one modality to another in a bi-directional manner, serving as a cross-modality pseudo-label generation process.
This module effectively assigns modality-unified pseudo-labels for individual instances, enabling the interaction between cross-modality instances. 
Afterward, leveraging the results from the dual association, we develop contrastive learning mechanisms based on global features and cluster-level memory banks.
Based on our discovery of inconsistency feature/pseudo-label distributions across cross-modality images, we propose a Fine-Grained Semantic-Aligned Learning (FGSAL) module.
This module delves into modality-related semantic-aligned patterns presented in the cross-modality images.
Specifically,
the semantic-aligned patterns are derived from potential positive cross-modality pairs through a query-guided attention mechanism.
Consider the visible modality as an illustrative example.
We first derive a descriptive query from a visible image, which encapsulates its fine-grained, identity-discriminative information prioritized by the visible modality.  
Such a visible query is utilized to extract part features from potential positive cross-modality images by aggregating their pixel-level features.
The aggregated part features share the same semantics associated with the specific visible query, so-called semantic-aligned features, which are subsequently optimized through part-level contrastive learning guided by the visible label space.
Drawing from the above analysis, we incorporate an instance-adaptive query generation module and a query-guided attention module to uncover modality-related semantic-aligned patterns.
Unlike existing methods that treat each image as an independent instance \cite{PartMix, PPLR, PCB}, such a query-guided attention mechanism leverages inter-image interactions to facilitate the learning of modality-shared fine-grained features from pairwise images.
As training goes on, cross-modality semantic-aligned features are optimized with respect to pseudo-labels in their relevant label space, progressively learning modality-shared information from fine-grained part-level patterns and enhancing the quality of pseudo-labels in subsequent epochs.

Nevertheless, the pseudo-labels from dual association inevitably contain noisy labels.
Persistently training with such static supervision leads to overfitting incorrect labels.
To multigate this challenge, we further design a Global-Part Collaborative Refinement (GPCR) module, aimed at the online discovery of reliable positive samples, and the optimization of inter-instance relationships for both global and part features.
Such an online module serves as a refinement for incorrect structure information arising from noisy offline pseudo-labels.
A straightforward idea is directly searching for nearest neighbors as the positive samples.
However, such simplicity limits the model's learning to easy instances while overlooking hard positives.
Therefore, we design two strategies to mine the reliable positive sets for global and part features respectively: a cross-modality intersection strategy for global features and a mutual correction strategy for part features. 
The meticulously crafted mining strategies incorporate hard negative samples into training while ensuring the sufficiency of the positive sets.
Subsequently, instance-level contrastive learning is built on top of the positive sets and the instance memory banks to constrain the inter-instance relationships.
During training, the GPCR module facilitates the exploration of complicated instance-level associations while dynamically adjusting the incorrect cluster-level structures introduced by noisy pseudo-labels.


To further enhance the cross-modality association, we propose a Cross-Modality Feature Propagation (CMFP) module to incorporate neighborhood information in the embedding space, serving as an efficient re-ranking algorithm for post-processing.
The simple-yet-effective CMFP module can be integrated into both the training and the testing stages for more precise associations and retrieval.
Different from existing re-ranking methods \cite{k-reciprocal, H2H, SAAI} which operates on the final dist matrix, our CMFP directly aggregates neighboring instances at the feature level.
It is demonstrated as a more efficient and effective solution compared to existing cross-modality re-ranking technologies. 

The main contribution can be summarized as follows:

\begin{itemize}
    \item 
    We design a SALCR framework with optimization objectives for semantic-aligned patterns salient in each modality, leveraging the discovery of inconsistent cross-modality label distributions.
    To the best of our knowledge, this is the first work to achieve fine-grained interaction between cross-modality images in USL-VI-ReID through a query-guided attention mechanism.
    \item 
    We devise a Global-Part Collaborative Refinement (GPCR) module to 
    dynamically mine reliable positive samples for global and part features, and optimize the inter-instance relationships, which adjusts the incorrect structures arising from noisy offline pseudo-labels.
    \item 
    A Cross-Modality Feature Propagation (CMFP) module is further proposed as an efficient re-ranking to enhance both association and retrieval.
    \item  Extensive experiments on two mainstream VI-ReID benchmarks demonstrate the effectiveness of our framework, elucidating the power of fine-grained features in USL-VI-ReID.
\end{itemize}


\section{Related Work}

\subsection{Unsupervised single-modality person ReID}

Unsupervised single-modality person ReID aims to retrieve pedestrian images from visible cameras without annotations. Most existing methods \cite{MMT, ClusterContrast, PPLR, ISE, DCMIP} are cluster-based, which alternate between pseudo-label generation and model training. 
To handle the inevitable noisy pseudo-labels, some methods \cite{PPLR, MMT, O2CAP, RLCC} utilize feature space information for refinement. 
MMT \cite{MMT} designs a mutual-mean-teacher framework for online refinement based on self-consistency.
RLCC \cite{RLCC} introduces the cluster consensus for generating temporal consistency pseudo-labels across training iterations.
PPLR \cite{PPLR} proposes the cross-agreement score to integrate part-level relationships into the pseudo-labels. 
O2CAP \cite{O2CAP} extensively establishes camera-aware proxies and associates the proxies online to refine the offline labels. 
Another stream of methods \cite{SPCL, ClusterContrast, ISE, DCMIP} devise memory-based contrastive learning frameworks to promote intra-cluster learning.   
SPCL \cite{SPCL} constructs both instance-level and cluster-level memory banks with a self-paced strategy to generate progressively reliable pseudo-labels. 
Cluster-Contrast \cite{ClusterContrast} utilizes a unique prototype to represent a cluster for maintaining updating consistency.
To explore the contextual information within the embedding space, ISE \cite{ISE} generates extensive samples implicitly to fill the cluster boundary, while DCMIP \cite{DCMIP} manages discrepant cluster prototypes and multi-instance prototypes to excavate multifaceted intra-cluster information.  

However, when directly applied to cross-modality scenarios, the above methods encounter challenges 
due to the absence of cross-modality interaction learning.

\subsection{Supervised VI-ReID}

Supervised VI-ReID aims to retrieve pedestrian images from both visible and infrared cameras with human annotations. Existing methods \cite{AlignGAN, hi-cmd, CA, MPANet, MAUM} primarily concentrate on bridging the modality gap at the image level and the feature level. 
To align the modalities at the image level,
some methods \cite{cmGAN, hi-cmd, AlignGAN} transfer images from one modality to another based on Generative Adversarial Networks (GANs). However, the inevitable noise in generated images and the high time consumption for training GANs hinder the real application of these methods. 
Other methods \cite{CA, LTG, MCJA} design various data augmentations to serve as intermediate modalities.
CA \cite{CA} proposes a simple random channel augmentation to effectively bridge the cross-modality images. 
Another stream of methods develops network architectures and metric learning strategies to align different modalities at the feature level.
Modality compensation methods \cite{zhang2022fmcnet, IDKL} utilize a multi-stream backbone with modality disentanglement to extract both modality-shared and modality-specific features.
Then the modality-specific features serve as compensations for modality-shared features.
Part-based methods \cite{MPANet, CAL, SAAI, PartMix} design novel part-mining blocks based on the attention mechanism to explore part representations for diverse regions.
Frequency domain-based methods \cite{Freq1, Freq2} mines cross-modality nuances in frequency and phase based on the FFT algorithm \cite{FFT}.

The above methods achieve remarkable performance on VI-ReID datasets. However, these methods rely on manually annotated cross-modality associations, which are labor-intensive and always inaccessible in real scenarios.
Thus we investigate the more data-friendly unsupervised VI-ReID task.

\subsection{Unsupervised VI-ReID}

Unsupervised VI-ReID aims to excavate reliable cross-modality associations without human annotations. Existing methods \cite{ADCA, PGM, MBCCM, GUR, DCCL, MMM} mainly follow a two-stage pipeline: (a) generating intra-modality pseudo-labels separately in each modality; (b) establishing cross-modality associations based on the intra-modality pseudo-labels. H2H \cite{H2H} proposes a homogeneous-to-heterogeneous training strategy and an ISML loss to establish online cross-modality associations based on reliable positive pairs.
OTLA \cite{OTLA} introduces a cross-modality label assignment algorithm based on optimal transport to avoid biased associations.
ADCA \cite{ADCA} set a robust DCL baseline for USL-VI-ReID and aggregates cross-modality memories through pairwise instance similarities.
PGM \cite{PGM} and MBCCM \cite{MBCCM} formulate the cross-modality cluster-level associations as bipartite graph matching problems from a global perspective.
CCLNet \cite{CCLNet} first proposes to 
utilize the powerful semantic information from CLIP \cite{CLIP} as supervision for contrastive learning.
To discover multi-view information within clusters, MMM \cite{MMM} proposes a multi-memory matching for more precise associations, and PCMIP \cite{PCMIP} maintains multi-proxies for each cluster to promote cross-modality learning.
To mine unified representations for clustering, GUR \cite{GUR} further introduces a CAE module to embed hierarchical domain information and achieves impressive performance.

Nevertheless, existing methods mainly focus on learning modality-invariant global features. 
We claim that relying solely on global features for cross-modality interaction is insufficient for learning robust representations.
Therefore, we blend the fine-grained semantic-aligned features into cross-modality interactions from a novel perspective of leveraging the complementarity between inconsistent cross-modality label distributions.

\section{Proposed Method}

\begin{figure*}[h]
\centering
\includegraphics[width=1.0\textwidth]{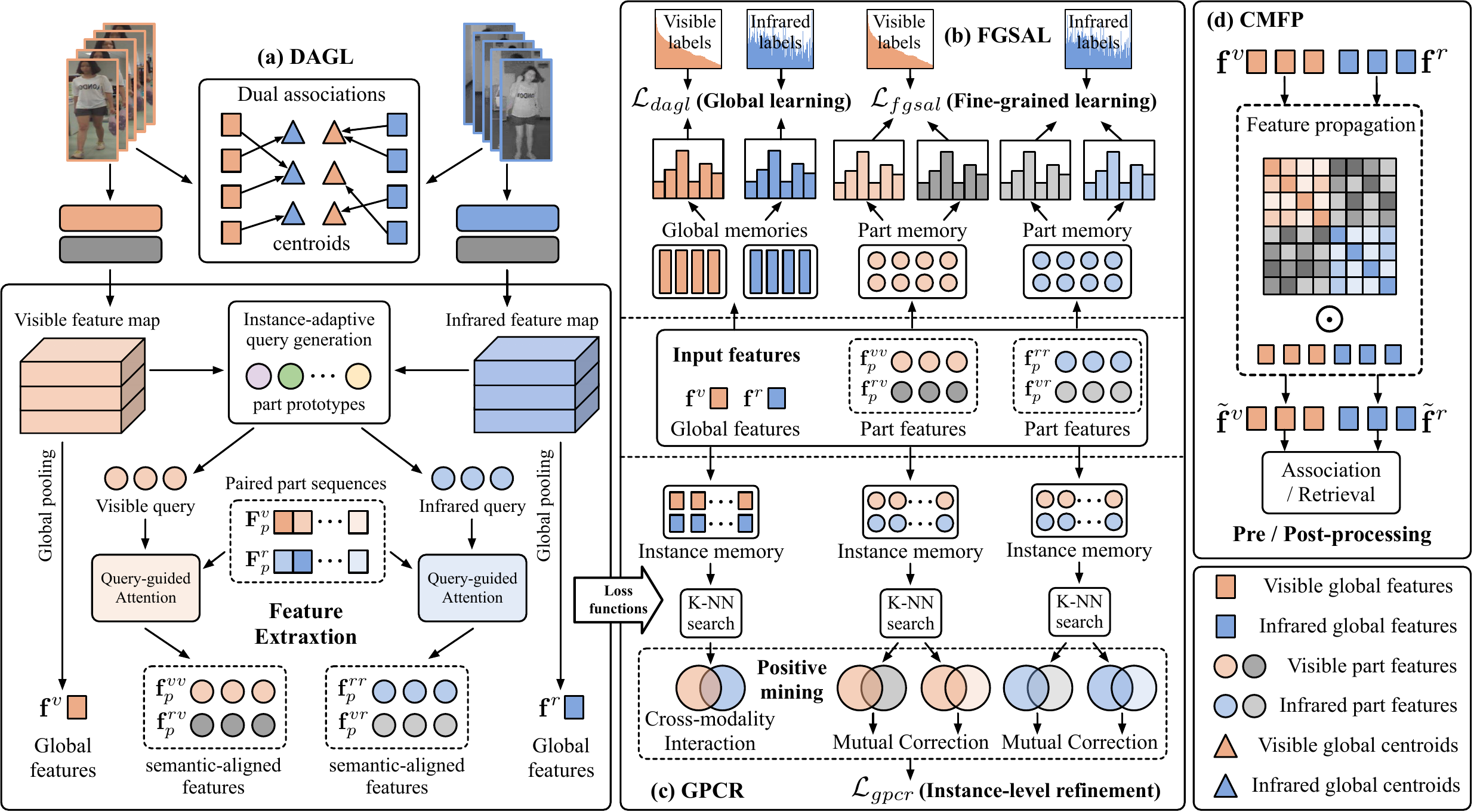}
\vspace{-3mm}
\caption{Overall framework of our proposed SALCR. Our framework mainly contains four components: (a) Dual Association with Global Learning (DAGL); (b) Fine-Grained Semantic-Aligned Learning (FGSAL); (c) Global-Part Collaborative Refinement (GPCR) and (d) Cross-Modality Feature Propagation (CMFP). 
The DAGL module generates modality-unified pseudo-labels at the beginning of each epoch.
During training, the FGSAL and GPCR modules are executed.
The FGSAL module first explores fine-grained semantic-aligned patterns and then optimizes them through cluster-level memories.
The GPCR module mines reliable positive samples for global and part features from instance memories.
The CMFP module further enhances the association and retrieval as pre-processing and post-processing steps.
}\label{fig:framework}
\vspace{-5mm}
\end{figure*}

\textbf{Problem Definition.} Let $\mathcal{X}=\{\mathcal{X}^v,\mathcal{X}^r\}$ denote a visible-infrared dataset, where $\mathcal{X}^v = \{\mathbf{x}^v_i | i=1, 2, \ldots, N^v\}$ represents the visible dataset consisting of $N^v$ visible images, and $\mathcal{X}^r = \{\mathbf{x}^r_i | i=1, 2, \ldots, N^r\}$ represents the infrared dataset consisting of $N^r$ infrared images.
Our backbone model is a convolutional neural network (CNN) $f_{\theta}(\cdot)$, parameterized by $\theta$. The objective is to train the model to project an image $\mathbf{x}_i$ from the dataset $\mathcal{X}$ into a d-dimensional feature space, producing an identity-discriminative and modality-invariant feature $\mathbf{f}_i = f_{\theta}(\mathbf{x}_i) \in \mathbb{R}^d$ without any manual annotations.

\noindent \textbf{Overall Framework.} 
To tackle the challenge of USL-VI-ReID, we propose a Semantic-Aligned Learning with Collaborative Refinement (SALCR) framework, as illustrated in Fig.\ref{fig:framework}. 
Our framework adheres to the commonly used unsupervised pipeline, alternating between (1) pseudo-label generation and (2) network training.
During pseudo-label generation, we initially generate intra-modality pseudo-labels through clustering. 
Subsequently, we introduce the Dual Association with Global Learning (DAGL, Sec.\ref{DAGI}) module to establish bi-directional cross-modality associations, enabling representation learning guided by modality-unified labels.
During network training, a Fine-Grained Semantic-Aligned Learning module (FGSAL, Sec.\ref{PACL}) is proposed for optimization objective of fine-grained semantic-aligned patterns from pairwise cross-modality instances.
, which exploits the complementarity between pseudo-label spaces.
To alleviate the side-effects brought by noisy labels during training, we design a Global-Part Collaborative Refinement module (GPCR, Sec.\ref{GPCR}), which dynamically mines reliable positive samples for input features and serves as a refinement for incorrect offline pseudo-labels.
The Cross-Modality Feature Propagation module (CMFP, Sec.\ref{CMFP}) is executed as an efficient re-ranking algorithm to further enhance the association and retrieval processes.
Our baseline is a simple framework (Sec.\ref{baseline}) that only focuses on intra-modality pseudo-label generation and learning.


\subsection{Preliminaries}\label{baseline}


We first set up a straightforward baseline with intra-modality pseudo-label generation and contrastive learning.
At each epoch, we employ a two-stream encoder $f_{\theta}$ ($i.e.$, ResNet-50) as our backbone to extract image features $\{\mathbf{f}^v_i | i=1, 2, \ldots, N^v\}$ and $
\{\mathbf{f}^r_i | i=1, 2, \ldots, N^r\}$
from two modalities.
we adopt DBSCAN \cite{1996DBSCAN} to cluster image features within each modality, respectively. The isolated outliers are discarded according to the cluster results. Then we obtain datasets from both two modalities with their intra-modality pseudo-labels $\mathcal{X}^v = \{(\mathbf{x}^v_i, \tilde{y}^v_i) | i=1, 2, \ldots, N^v\}$ and $\mathcal{X}^r = \{(\mathbf{x}^r_i, \tilde{y}^r_i) | i=1, 2, \ldots, N^r\}$.
The corresponding cluster centroids are obtained as $\{\mathbf{C}^v_i\}_{i=1}^{K^v}$ and $\{\mathbf{C}^r_i\}_{i=1}^{K^r}$ by averaging the features within one cluster. 
$K^v$ and $K^r$ denote the number of clusters within visible and infrared modalities, respectively. 

At the beginning of each epoch, two cluster-level memory banks $\mathbf{M}^v \in \mathbb{R}^{K^v \times d}$ and $\mathbf{M}^r \in \mathbb{R}^{K^r \times d}$ are initialized by the cluster centroids $\{\mathbf{C}^v_i\}_{i=1}^{K^v}$ and $\{\mathbf{C}^r_i\}_{i=1}^{K^r}$, which store a unique prototype for each cluster.
During model training, the InfoNCE loss \cite{moco} is executed to perform intra-modality contrastive learning:

\begin{equation}\label{L_IM_v}
\mathcal{L}_{IM}^v = -\frac{1}{B} \sum_{i=1}^B
\log 
\frac{\exp(\mathbf{M}^v[\tilde{y}^v_i]^{\top} \cdot f_{\theta}(\mathbf{x}_i^v) / \tau)}{\sum_{j=1}^{K^v} \exp(\mathbf{M}^v[j]^{\top} \cdot f_{\theta}(\mathbf{x}_i^v) / \tau)},  
\end{equation}

\vspace{-1mm}
\begin{equation}\label{L_IM_r}
\mathcal{L}_{IM}^r = -\frac{1}{B} \sum_{i=1}^B
\log 
\frac{\exp(\mathbf{M}^r[\tilde{y}^r_i]^{\top} \cdot f_{\theta}(\mathbf{x}_i^r) / \tau)}{\sum_{j=1}^{K^r} \exp(\mathbf{M}^r[j]^{\top} \cdot f_{\theta}(\mathbf{x}_i^r) / \tau)},  
\end{equation}

\noindent where $B$ is the batch size and $\tau$ is a temperature parameter. $\mathbf{x}_i^v$ and $\mathbf{x}_i^r$ denote the input visible and infrared images. $\mathbf{M}^v[\tilde{y}^v_i]$ denotes the $\tilde{y}^v_i$-th prototype in memory bank $\mathbf{M}^v$. 
The intra-modality loss for our baseline is formulated as follows:

\begin{equation}
    \mathcal{L}_{intra} = \mathcal{L}_{IM}^v+ 
    \mathcal{L}_{IM}^r.
\end{equation}

Following the DCL framework \cite{ADCA}, we also adopt the random channel augmentation \cite{CA} for visible images as an augmented branch.

\subsection{Dual Association with Global Learning}\label{DAGI}


To guide the cross-modality interaction with modality-unified labels, we first design a dual association algorithm to assign pseudo-labels in one modality to another in a bi-directional manner.
Without loss of generality, we take the infrared instances as an example.
To assign pseudo-labels of visible clusters to infrared instances, we inject an OTLA-base association into our memory-based framework.
The Optimal Transport Label Assignment (OTLA) \cite{OTLA} has demonstrated an effective approach for cross-modality label association in the Semi-Supervised VI-ReID task \cite{OTLA, DPIS}.
In our memory-based framework, we leverage the cross-modality instance-cluster relationships to construct OTLA, which is formulated as follows:


\begin{equation}\label{eq:OTLA}
\begin{aligned}
    &\mathop{\min} \limits_{\mathbf{Q}} \langle \mathbf{Q}, \mathbf{P} \rangle + \frac{1}{\lambda_{ot}} \langle \mathbf{Q}, -\log(\mathbf{Q}) \rangle. \\
    s.t. &\quad \mathbf{Q} \mathds{1} = \mathds{1} \cdot \frac{1}{N^v}, \quad \mathbf{Q}^\top \mathds{1} = \mathds{1} \cdot \frac{1}{K^r},
\end{aligned}
\end{equation}

\noindent where $\langle \cdot \rangle$
denotes the Frobenius dot-product,
and $\mathds{1}$ is an all in 1 vector.
$\mathbf{Q} \in \mathbb{R}^{N^r \times K^v}$ is the transport plan and $\mathbf{P} \in \mathbb{R}^{N^r \times K^v}$ is the cost matrix. 
The Euclidean distance between the infrared instances and the visible prototypes in feature space is utilized to obtain the cost matrix, where $\mathbf{P}_{ij} = \Vert f_{\theta}(\mathbf{x}_i^r) - \mathbf{M}^v[j] \Vert_2^2$. 
The optimal solution $\mathbf{Q}^{\ast}$ of Eq.\ref{eq:OTLA} can be solved by the Sinkhorn-Knopp algorithm \cite{sinkhorn}.
Then we derive the visible cluster labels assigned to infrared instances from $\mathbf{Q}^{\ast}$, denoted as $\{\hat{y}^r_i\}_{i=1}^{N^r}$, where $\hat{y}^r_i = \mathop{\arg\max}_j (\mathbf{Q}^{\ast}_{ij})$.
Therefore, $i$-th infrared instance holds two types of pseudo-labels $\tilde{y}^r_i$ and $\hat{y}^r_i$ in two label spaces.
The infrared cluster labels $\{\hat{y}^v_i\}_{i=1}^{N^v}$ of visible instances can be derived similarly. 
The above two pseudo-label sets are utilized for contrastive learning to learn modality-invariant representations:

\vspace{-2mm}
\begin{equation}\label{L_CM_v}
\mathcal{L}_{CM}^v = -\frac{1}{B} \sum_{i=1}^B
\log 
\frac{\exp(\mathbf{M}^v[\hat{y}^r_i]^{\top} \cdot f_{\theta}(\mathbf{x}_i^r) / \tau)}{\sum_{j=1}^{K^v} \exp(\mathbf{M}^v[j]^{\top} \cdot f_{\theta}(\mathbf{x}_i^r) / \tau)},  
\end{equation}
\vspace{-2mm}

\begin{equation}\label{L_CM_r}
\mathcal{L}_{CM}^r = -\frac{1}{B} \sum_{i=1}^B
\log 
\frac{\exp(\mathbf{M}^r[\hat{y}^v_i]^{\top} \cdot f_{\theta}(\mathbf{x}_i^v) / \tau)}{\sum_{j=1}^{K^r} \exp(\mathbf{M}^r[j]^{\top} 
\cdot f_{\theta}(\mathbf{x}_i^v) / \tau)}.  
\end{equation}


\noindent Following existing memory-based methods \cite{SPCL, ClusterContrast, ADCA, MBCCM, PGM}, the memory banks $\mathbf{M} = \{\mathbf{M}^v$ $\mathbf{M}^r\}$ are updated during back-propagation (BP) with a momentum strategy:

\begin{equation}\label{eq:update_momentum}
    \mathbf{M}[y] \leftarrow \mu \mathbf{M}[y] + (1 - \mu) \mathbf{f}[y],
\end{equation}

\noindent where $\mathbf{f}[y]$ denotes the input feature 
$\mathbf{f} \in \{\mathbf{f}^v_i\}_{i=1}^B \cup \{\mathbf{f}^r_i\}_{i=1}^B$ with its corresponding label $y$ within a mini-batch and $\mu$ is the momentum updating factor.
The total loss functions for our DAGL module can be formulated as follows:

\begin{equation}\label{eq:total_base}
    \mathcal{L}_{dagl} = 
    \mathcal{L}_{intra} +
    \mathcal{L}_{CM}^v + \mathcal{L}_{CM}^r.
\end{equation}

\subsection{Fine-Grained Semantic-Aligned Learning}\label{PACL}

The FGSAL module aims to explore fine-grained semantic-aligned features emphasized by each modality and jointly optimize them through two label spaces. 
This module first generates instance-adaptive queries that capture the identity-discriminative and modality-aware information corresponding to each instance part. 
Then the queries are utilized in a query-guided attention mechanism to mine semantic-aligned features from cross-modality instances. 
Finally, part-level modality-aware contrastive learning is carried out to constrain the fine-grained features in the embedding space.

To generate instance-adaptive queries, 
it first set up $N_p$ learnable modality-shared prototypes $\mathbf{p}_s = [\mathbf{p}_1, \mathbf{p}_2, \cdots, \mathbf{p}_{N_p}] \in \mathbb{R}^{N_p \times d}$, where each prototype contains shared information for a unique part. 
For visible modality, we first derive the feature maps output from the 4-th layer of the backbone ($i.e.$, ResNet-50).
The flattened feature map before the global pooling operation is denoted as
$\mathbf{F}^v = [\mathbf{F}^v_1, \mathbf{F}^v_2, \cdots, \mathbf{F}^v_{HW}] \in \mathbb{R}^{HW \times d}$, where $H$ and $W$ are the height and width of the feature map.
Then the feature map is divided into $N_p$ sequences
$\{\mathbf{F}^v_p\}_{p=1}^{N_p}$, where $\mathbf{F}^v_{p} = [\mathbf{F}^v_{\lceil \frac{HW}{N_p} (p-1) \rceil}, \cdots, \mathbf{F}^v_{\lceil \frac{HW}{N_p} p \rceil}] \in \mathbb{R}^{m_p \times d}$, $m_p$ is the sequence length and $\lceil \cdot \rceil$ denotes the ceiling operation.
Each sequence contains all pixel information corresponding to the respective part. 
The query for $p$-th part can be obtained as a weighted sum of the pixel features within the $p$-th sequence $\mathbf{F}^v_p$.
To derive the instance-adaptive query, we aggregate pixel-level features according to their similarities with the corresponding part prototype:

\begin{equation}\label{eq:query}
    \mathbf{A} = \sigma 
    (\mathbf{p}_p \cdot {\mathbf{F}^v_{p}}^{\top}) \in \mathbb{R}^{1 \times m_p}, 
    \mathbf{q}^v_p = \mathbf{A} \cdot {\mathbf{F}^v_{p}}^{\top} / \sum_{i,j} \mathbf{A}_{ij},
\end{equation}

\begin{figure}[t]
\vspace{-1mm}
\centering
\includegraphics[width=0.48\textwidth]{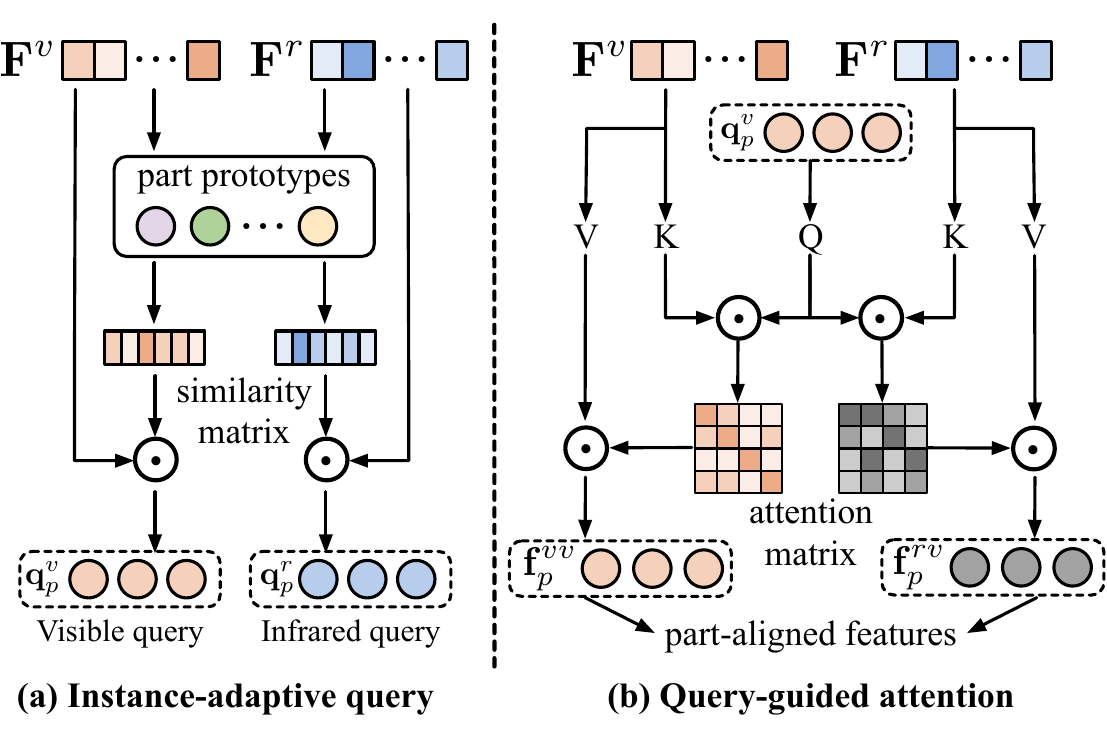}
\caption{Illustration of (a) the instance-adaptive query generation module and (b) query-guided attention.}\label{fig:PACL}
\vspace{-3mm}
\end{figure}

\noindent where $\sigma$ denotes the Sigmoid activation function and $\mathbf{A}$ denotes the similarity matrix between pixel-level features in $\mathbf{F}^v_p$ and $p$-th prototype $\mathbf{p}_p$. 
The query $\mathbf{q}^r_p$ for infrared feature map $\mathbf{F}^r_p$ can be obtained similarly. 
Fig.\ref{fig:PACL}(a) illustrates the instance-adaptive query generation module.
The instance-adaptive queries are subsequently utilized to explore semantic-aligned features.

To explore semantic-aligned features from pairwise cross-modality instances, we adopt the Scaled Dot-Product Attention \cite{Attention} between a specific instance-adaptive query and a potential positive cross-modality sequence pair to excavate the query-associated patterns.
Suppose two $p$-th part sequence, $\mathbf{F}^v_p$ and $\mathbf{F}^r_p$ from visible and infrared modalities, respectively.
A potential positive pair is defined when they hold the shared pseudo-label in the label space of either modality, where $\tilde{y}^v[\mathbf{F}^v_p] = \hat{y}^r[\mathbf{F}^r_p]$ or $\hat{y}^v[\mathbf{F}^v_p] = \tilde{y}^r[\mathbf{F}^r_p]$.
$y[\mathbf{F}_p]$ denotes the pseudo-label $y$ corresponding to $\mathbf{F}_p$. 
We take the part query $\mathbf{q}^v_p$ generated from $\mathbf{F}^v_p$ as $Q$, the part sequence $\mathbf{F}^v_p$ or $\mathbf{F}^r_p$ as both $K$ and $V$ to obtain a pair of the query-associated part features:

\vspace{0mm}
\begin{equation}\label{eq:PA1}
\begin{aligned}
    \mathbf{A}^{vv}_p = {\rm Softmax} (\mathbf{q}^v_p \cdot {\mathbf{F}^v_p}^{\top} / \sqrt{d}) \in \mathbb{R}^{1 \times m_p}, \\
    \mathbf{A}^{rv}_p = {\rm Softmax} (\mathbf{q}^v_p \cdot {\mathbf{F}^r_p}^{\top} / \sqrt{d}) \in \mathbb{R}^{1 \times m_p},
\end{aligned}
\end{equation}

\vspace{-2mm}
\begin{equation}\label{eq:PA2}
    \mathbf{f}^{vv}_p = \mathbf{A}^{vv}_p \cdot {\mathbf{F}^v_p}^{\top} \in \mathbb{R}^{d}, 
    \mathbf{f}^{vr}_p = \mathbf{A}^{vr}_p \cdot {\mathbf{F}^r_p}^{\top} \in \mathbb{R}^{d},
\end{equation}

\noindent where $\mathbf{A}^{vv}_p$ and $\mathbf{A}^{vr}_p$ denote the attention maps. 
Then we derive a pair of aligned series of part features $\{\mathbf{f}^{vv}_p\}_{p=1}^{N_p}$ and $\{\mathbf{f}^{rv}_p\}_{p=1}^{N_p}$ from the same query set $\{\mathbf{q}^v_p\}_{p=1}^{N_p}$ in visible modality.
Another pair $\{\mathbf{f}^{rr}_p\}_{p=1}^{N_p}$ and $\{\mathbf{f}^{rv}_p\}_{p=1}^{N_p}$ can be obtained in the same manner from $\{\mathbf{q}^r_p\}_{p=1}^{N_p}$. 
A pair of part features $\mathbf{f}^{vv}_p$ and $\mathbf{f}^{rv}_p$ from $\mathbf{q}^v_p$ means they are both reorganized from the original pixels through the perspective of visible query $\mathbf{q}^v_p$, thus containing information highlighted by the visible modality.
Furthermore, they integrate the pixels containing semantic information relevant to the specific query, so-called semantic-aligned features. 
These paired part features are subsequently optimized utilizing the pseudo-labels in their corresponding label space through part-level contrastive learning.

At the beginning of each epoch, for $p$-th part,
two modality-aware part-level memory banks $\tilde{\mathbf{M}}^v_p \in \mathbb{R}^{K^v \times d}$ and 
$\hat{\mathbf{M}}^r_p \in \mathbb{R}^{K^v \times d}$ are initialized by the cluster centroids of part features 
$\mathbf{f}^{vv}_p$ based on pseudo-labels $\tilde{y}^v$.
Similarly, another two memory banks $\hat{\mathbf{M}}^v_p \in \mathbb{R}^{K^r \times d}$ and 
$\tilde{\mathbf{M}}^r_p \in \mathbb{R}^{K^r \times d}$ are initialized by the cluster centroids of part features 
$\mathbf{f}^{rr}_p$ based on pseudo-labels $\tilde{y}^r$.
The memory banks $\tilde{\mathbf{M}}^v_p$ and $\hat{\mathbf{M}}^v_p$ store the part features $\{\mathbf{f}^{vv}_p, \mathbf{f}^{rv}_p\}$ aggregated from visible queries, and vice versa.
During training, a pair of semantic-aligned features $\mathbf{f}^{vv}_p$ and $\mathbf{f}^{rv}_p$, associated with the query $\mathbf{q}^v_p$, are supervised by the shared pseudo-labels of $\mathbf{F}^v_p$ in the visible label space ($i.e.$, $\tilde{y}^v[\mathbf{F}^v_p]$ and $\hat{y}^v[\mathbf{F}^v_p]$).
The contrastive loss for $\mathbf{f}^{vv}_p$ can be formulated as follows:

\vspace{-2mm}
\begin{equation}\label{L^vv1(p)_P}
    \mathcal{L}_P^{vv1(p)} = 
    -\frac{1}{| \mathcal{Q}_{vv} |} \sum_{\mathbf{f}^{vv}_p \in \mathcal{Q}_{vv}} \log \frac{\exp(\tilde{\mathbf{M}}^v_p[\tilde{y}^v[\mathbf{F}^v_p]] \cdot {\mathbf{f}^{vv}_p}^{\top} / \tau)}{\sum_{j=1}^{K^v} \exp(\tilde{\mathbf{M}}^v_p[j] 
    \cdot {\mathbf{f}^{vv}_p}^{\top} / \tau)},
\end{equation}

\vspace{-6mm}
\begin{equation}\label{L^vv2(p)_P}
    \mathcal{L}_P^{vv2(p)} = 
    -\frac{1}{| \mathcal{Q}_{vv} |} \sum_{\mathbf{f}^{vv}_p \in \mathcal{Q}_{vv}} \log \frac{\exp(\hat{\mathbf{M}}^v_p[\hat{y}^v[\mathbf{F}^v_p]] \cdot {\mathbf{f}^{vv}_p}^{\top} / \tau)}{\sum_{j=1}^{K^v} \exp(\hat{\mathbf{M}}^v_p[j] 
    \cdot {\mathbf{f}^{vv}_p}^{\top} / \tau)},
\end{equation}

\noindent where $\mathcal{Q}_{vv}$ is the part feature set containing all $\mathbf{f}^{vv}_p$ within a mini-batch, and $| \mathcal{Q}_{vv} |$ is the cardinality of $\mathcal{Q}_{vv}$. 
The contrastive loss $\mathcal{L}_P^{rv1(p)}$ and $\mathcal{L}_P^{rv2(p)}$ for $\mathbf{f}^{rv}_p$ can be formulated in the same manner. 
The total part-level contrastive loss for pairwise semantic-aligned features $\mathbf{f}^{vv}_p$ and $\mathbf{f}^{rv}_p$ is obtained as the sum of the above four terms:




\begin{equation}
    \mathcal{L}_P^{v(p)} = \mathcal{L}_P^{vv1(p)} + \mathcal{L}_P^{vv2(p)} + \mathcal{L}_P^{rv1(p)} + \mathcal{L}_P^{rv2(p)}.
\end{equation}

Through the above loss terms, 
$\tilde{y}^v[\mathbf{F}^v_p]$ and $\hat{y}^v[\mathbf{F}^v_p]$ in the visible label space focus on optimizing the specific patterns emphasized by the visible modality of instances in two modalities, thus achieving targeted fine-grained learning.
The contrastive loss $\mathcal{L}_P^{r(p)}$ for another pair of semantic-aligned features $\mathbf{f}^{rv}_p$ and $\mathbf{f}^{rr}_p$ emphasized by the infrared modality can be derived similarly based on pseudo-labels $\tilde{y}^r$ and $\hat{y}^r$, and the corresponding memory banks $\hat{\mathbf{M}}^r_p$ and $\tilde{\mathbf{M}}^r_p$.
The total loss for the FGSAL module is formulated as follows:

\begin{equation}\label{eq:total_pacl}
    \mathcal{L}_{fgsal} = 
    \frac{1}{N_p} \sum_{p=1}^{N_p} (\mathcal{L}_P^{v(p)} + \mathcal{L}_P^{r(p)}).
\end{equation}

\noindent During the backward propagation, the part-level memory banks are updated by the input part features in the same manner as Eq.\ref{eq:update_momentum}.

\subsection{Global-Part Collaborative Refinement}\label{GPCR}


To alleviate the adverse effects of noisy pseudo-labels,
we propose GPCR, which dynamically discovers reliable positive samples of both global and part features and builds up instance-level contrastive learning between input features and positive samples.
We devise a cross-modality strategy and a mutual correction strategy for global features and semantic-aligned part features, respectively.
Such a dynamic module adjusts the incorrect cluster-level structure introduced by noisy pseudo-labels, thus functioning as an online refinement for offline pseudo-labels.

\begin{figure}[t]
\centering
\includegraphics[width=0.48\textwidth]{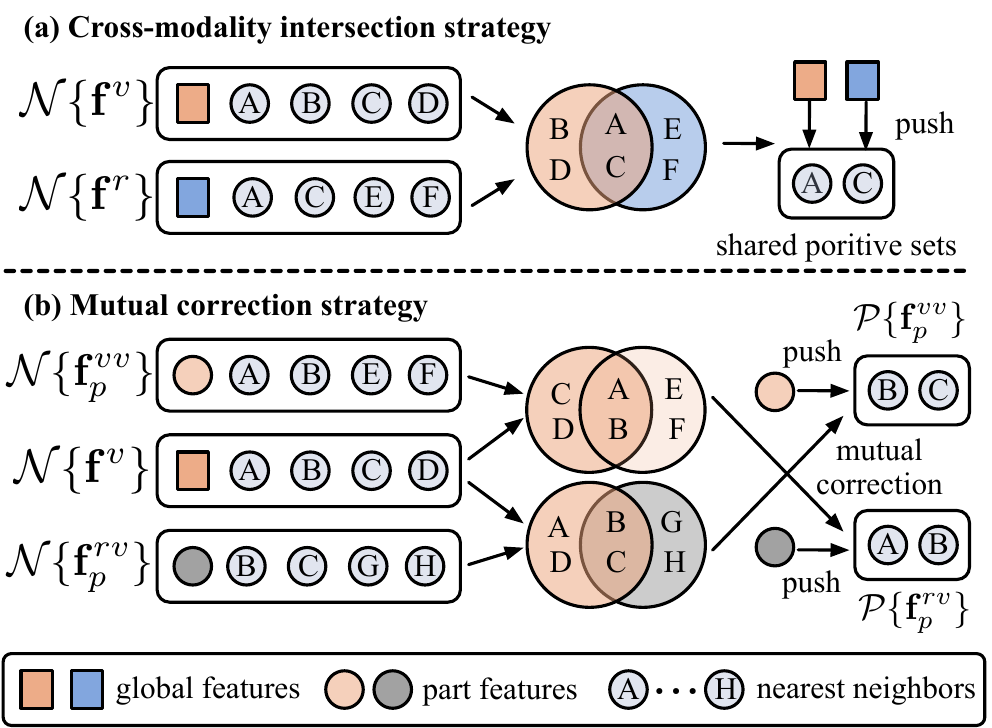}
\caption{Illustration of the cross-modality intersection strategy (a) for global features and the mutual correction strategy (b) for part features. 
Circles with letters represent k-nearest neighbors of the input features.
Different letters indicate different instance indexes in instance memory banks.
}\label{fig:GPCR}
\vspace{-5mm}
\end{figure}

Specifically, at each epoch, two global-level instance memory banks, denoted as $\mathbf{M}^v_I \in \mathbb{R}^{N^v \times d}$ and $\mathbf{M}^r_I \in \mathbb{R}^{N^r \times d}$, are initialized by global features from two modalities. 
Additionally, two part-level instance memory banks $\mathbf{M}^{vp}_I \in \mathbb{R}^{N^v \times d}$ and $\mathbf{M}^{rp}_I \in \mathbb{R}^{N^r \times d}$ for $p$-th part are initialized by $\mathbf{f}^{vv}_p$ and $\mathbf{f}^{rr}_p$ extracted from the trainset. 
During training, a potential positive pair with their global features $\mathbf{f}^v$ and $\mathbf{f}^r$, and their $p$-th semantic-aligned part features $\{\mathbf{f}^{vv}_p, \mathbf{f}^{rv}_p\}, \{\mathbf{f}^{vr}_p, \mathbf{f}^{rr}_p\}$ serve as input to the GPCR module.
We first introduce a cross-modality intersection strategy for searching reliable positive samples of global features.
As illustrated in Fig.\ref{fig:GPCR}(a),
for global feature pairs $\mathbf{f}^v$ and $\mathbf{f}^r$, the intersection of their $k$-nearest samples in memory banks $\mathbf{M}^v_I$ and $\mathbf{M}^r_I$ is regarded as their shared positive sample set:

\vspace{-3mm}
\begin{equation}
    \mathcal{P}^{v} \{\mathbf{f}^v\} =   
    \mathcal{P}^{v} \{\mathbf{f}^r\} =
    \mathcal{N} \{\mathbf{f}^v, \mathbf{M}^v_I, k\} \cap 
    \mathcal{N} \{\mathbf{f}^r, \mathbf{M}^v_I, k\},
\end{equation}

\vspace{-6mm}
\begin{equation}
    \mathcal{P}^{r} \{\mathbf{f}^v\} = 
    \mathcal{P}^{r} \{\mathbf{f}^r\} =
    \mathcal{N} \{\mathbf{f}^v, \mathbf{M}^r_I, k\} \cap 
    \mathcal{N} \{\mathbf{f}^r, \mathbf{M}^r_I, k\},
\end{equation}

\noindent where $\mathcal{N} \{\mathbf{f}^v, \mathbf{M}^v_I, k\}$ denotes the index set of the $k$-nearest neighbor samples of $\mathbf{f}^v$ in memory $\mathbf{M}^v_I$. For visible global feature $\mathbf{f}^v$, the instance-level contrastive learning loss is formulated as follows:

\vspace{-3mm}
\begin{equation}
    \mathcal{L}^{vg1}_I = -\frac{1}{|\mathcal{Q}_{v}|} \sum_{\mathbf{f}^v \in \mathcal{Q}_{v}} \log 
    \frac{
    \sum_{i \in \mathcal{P}^{v} \{\mathbf{f}^v\}} 
    {\exp(\mathbf{M}^v_I[i] \cdot {\mathbf{f}^v}^{\top} / \tau)}}
    {\sum_{j=1}^{N^v} \exp(\mathbf{M}^v_I[j] \cdot {\mathbf{f}^v}^{\top} / \tau)},
\end{equation}

\vspace{-3mm}
\begin{equation}
    \mathcal{L}^{vg2}_I = -\frac{1}{|\mathcal{Q}_{v}|} \sum_{\mathbf{f}^v \in \mathcal{Q}_{v}} \log 
    \frac{
    \sum_{i \in \mathcal{P}^{r} \{\mathbf{f}^v\}} 
    {\exp(\mathbf{M}^r_I[i] \cdot {\mathbf{f}^v}^{\top} / \tau)}}
    {\sum_{j=1}^{N^r} \exp(\mathbf{M}^r_I[j] \cdot {\mathbf{f}^v}^{\top} / \tau)},
\end{equation}

\noindent where $\mathcal{Q}_{v}$ is the global feature set within a mini-batch. 
The contrastive loss $\mathcal{L}^{rg1}_I$ and $\mathcal{L}^{rg2}_I$ for infrared global feature $\mathbf{f}^r$ can be formulated in a similar manner. The total loss for global features in GPCR is formulated as follows:

\begin{equation}
    \mathcal{L}_I^g = \mathcal{L}^{vg1}_I + \mathcal{L}^{vg2}_I + \mathcal{L}^{rg1}_I + \mathcal{L}^{rg2}_I.
\end{equation}

We subsequently design a mutual correction strategy for part features.
We claim that the cross-modality semantic-aligned features $\mathbf{f}^{vv}_p$ and $\mathbf{f}^{rv}_p$ associated with a specific query $\mathbf{q}^v_p$ should encapsulate the modality-shared information with the same semantics.
Therefore, their corresponding positive sample sets can be interchanged.
Consequently, we formulate a mutual correction object between pairwise semantic-aligned features,
in which the neighbor set of $\mathbf{f}^{rv}_p$ is utilized as the positive set of $\mathbf{f}^{vv}_p$, and vice versa. 
However, the neighbor set of the part features is not reliable enough in the initial training stage.
To overcome this problem, we take their intersection with the neighbor sets of global features to mine robust positive gallery sets, as shown in Fig.\ref{fig:GPCR}(b). The positive sets for $\mathbf{f}^{vv}_p$ and $\mathbf{f}^{rv}_p$ within $\mathbf{M}^{vp}_I$ can be formulated as follows:

\vspace{-2mm}
\begin{equation}\label{eq:P_vv}
    \mathcal{P}^{v} \{\mathbf{f}^{vv}_p\}= 
    \mathcal{N} \{\mathbf{f}^v, \mathbf{M}^v_I, k\} \cap 
    \mathcal{N} \{\mathbf{f}^{rv}_p, \mathbf{M}^{vp}_I, k\},
\end{equation}

\vspace{-4mm}
\begin{equation}\label{eq:P_rv}
    \mathcal{P}^{v} \{\mathbf{f}^{rv}_p\}= 
    \mathcal{N} \{\mathbf{f}^v, \mathbf{M}^v_I, k\} \cap 
    \mathcal{N} \{\mathbf{f}^{vv}_p, \mathbf{M}^{vp}_I, k\}.
\end{equation}

\noindent 
Following DCL framework \cite{ADCA}, for each visible image, we take its original form $\mathbf{x}^v$ and its channel augmented \cite{CA} form $\mathbf{x}^a$ as inputs in a mini-batch. 
Therefore, we derive a triplet of semantic-aligned features $\{\mathbf{f}^{vv}_p, \mathbf{f}^{av}_p, \mathbf{f}^{rv}_p\}$ from query $\mathbf{q}^v_p$.
$\mathbf{f}^{av}_p$ denotes the part feature from $\mathbf{x}^a$ by taking $\mathbf{q}^v_p$ as query.
We expand Eq.\ref{eq:P_vv} by the neighbor sets within a triplet to explore additional potential positive instances for contrastive learning:

\begin{equation}\label{eq:P_vv}
\begin{aligned}
    &\mathcal{P}^{v} \{\mathbf{f}^{vv}_p\}= 
    \mathcal{N} \{\mathbf{f}^v, \mathbf{M}^v_I, k\} \cap \\
    &(\mathcal{N} \{\mathbf{f}^{av}_p, \mathbf{M}^{vp}_I, k\}
    \cup \mathcal{N} \{\mathbf{f}^{rv}_p, \mathbf{M}^{vp}_I, k\}),
\end{aligned}
\end{equation}

\noindent 
The positive set of $\mathbf{f}^{vv}_p$ is determined by the neighbor set of $\mathbf{f}^{av}_p$ and $\mathbf{f}^{rv}_p$ jointly.
Such a design avoids overfitting to hard negatives while ensuring the sufficiency of the positive samples.
Eq.\ref{eq:P_rv} can be written in an improved form similarly.
The corresponding part-level contrastive loss can be formulated according to the positive sets:

\vspace{-3mm}
\begin{equation}\label{eq:L_vpi}
\begin{aligned}
& \mathcal{L}^{vv1(p)}_I =\\
& -\frac{1}{|\mathcal{Q}_{vv}|} \sum_{\mathbf{f}^{vv}_p \in \mathcal{Q}_{vv}} \log
\frac{
\sum_{i \in \mathcal{P}^{v} \{\mathbf{f}^{vv}_p\}} 
{\exp(\mathbf{M}^{vp}_I[i] \cdot {\mathbf{f}^{vv}_p}^{\top} / \tau)}}
{\sum_{j=1}^{N^v} \exp(\mathbf{M}^{vp}_I[j] \cdot {\mathbf{f}^{vv}_p}^{\top} / \tau)},
\end{aligned}
\end{equation}

\vspace{-6mm}
\begin{equation}\label{eq:L_rpi}
\begin{aligned}
& \mathcal{L}^{rv1(p)}_I = \\
& -\frac{1}{|\mathcal{Q}_{rv}|} \sum_{\mathbf{f}^{rv}_p \in \mathcal{Q}_{rv}} \log 
\frac{
\sum_{i \in \mathcal{P}^{v} \{\mathbf{f}^{rv}_p\}} 
{\exp(\mathbf{M}^{vp}_I[i] \cdot {\mathbf{f}^{rv}_p}^{\top} / \tau)}}
{\sum_{j=1}^{N^v} \exp(\mathbf{M}^{vp}_I[j] \cdot {\mathbf{f}^{rv}_p}^{\top} / \tau)}.
\end{aligned}
\end{equation}

The positive sets $\mathcal{P}^{r} \{\mathbf{f}^{vv}_p\}$ and $\mathcal{P}^{r} \{\mathbf{f}^{rv}_p\}$ for $\mathbf{f}^{vv}_p$ and $\mathbf{f}^{rv}_p$ within $\mathbf{M}^{rp}_I$ can be obtained similarly. Then the contrastive loss $\mathcal{L}^{vv2(p)}_I$ and $\mathcal{L}^{rv2(p)}_I$ based on $\mathbf{M}^{rp}_I$ can be derived in the similar way as Eq.\ref{eq:L_vpi} and Eq.\ref{eq:L_rpi}. The total loss for semantic-aligned feature pair $\mathbf{f}^{vv}_p$ and $\mathbf{f}^{rv}_p$ can be formulated as:

\vspace{-1mm}
\begin{equation}\label{eq:L_IM_vp}
    \mathcal{L}_I^{v(p)} = \mathcal{L}^{vv1(p)}_I + \mathcal{L}^{rv1(p)}_I + 
    \mathcal{L}^{vv2(p)}_I + 
    \mathcal{L}^{rv2(p)}_I.
\end{equation}

The total loss $\mathcal{L}_I^{r(p)}$ of another semantic-aligned pair $\mathbf{f}^{vr}_p$ and $\mathbf{f}^{rr}_p$ holds a symmetric form of Eq.\ref{eq:L_IM_vp}. The total loss of the GPCR module is obtained as follows:

\vspace{-2mm}
\begin{equation}\label{eq:total_gpcr}
    \mathcal{L}_{gpcr} = 
    \mathcal{L}_I^g + \frac{1}{N_p}\sum_{p=1}^{N_p}
    (\mathcal{L}_I^{v(p)} + \mathcal{L}_I^{r(p)}).
\end{equation}
\vspace{-2mm}

\subsection{Cross-Modality Feature Propagation}\label{CMFP}


In this section, we introduce a simple yet effective CMFP algorithm to enhance the association and retrieval stages.
At each epoch, the normalized features $\{\mathbf{f}^v_i\}_{i=1}^{N^v}$ and $\{\mathbf{f}^r_i\}_{i=1}^{N^r}$ are extracted from the backbone. For visible modality, we construct both intra-modality and cross-modality similarity matrices $\mathbf{S}^{vv} \in \mathbb{R}^{N^v \times N^v}$ and $\mathbf{S}^{vr} \in \mathbb{R}^{N^v \times N^r}$ based on pairwise cosine similarities, where $\mathbf{S}^{vv}_{ij}=\mathbf{f}^v_i {\mathbf{f}^v_j}^{\top}$ and $\mathbf{S}^{vr}_{ij} =\mathbf{f}^v_i {\mathbf{f}^r_j}^{\top}$.
$\mathbf{S}^{rr}$ and $\mathbf{S}^{rv}$ are obtained similarly. 
We combine the above four matrices to construct the global affinity matrix:

\vspace{-2mm}
\begin{equation}\label{eq:G_tr}
\mathbf{G}_{tr} = 
\left[
\begin{array}{cc;{3pt/3pt}cc}
\multicolumn{2}{c;{3pt/3pt}}{\raisebox{0.7ex}[0pt]{$\overline{\mathcal{K}(\mathbf{S}^{vv}, k_{tr})}$}} & 
\multicolumn{2}{c}{\raisebox{0.7ex}[0pt]{$\overline{\mathcal{K}(\mathbf{S}^{vr}, k_{tr})}$}}\\
\hdashline[3pt/3pt]
\multicolumn{2}{c;{3pt/3pt}}{\raisebox{-1.3ex}[0pt]{$\overline{\mathcal{K}(\mathbf{S}^{rv}, k_{tr})}$}} & 
\multicolumn{2}{c}{\raisebox{-1.3ex}[0pt]{$\overline{\mathcal{K}(\mathbf{S}^{rr}, k_{tr})}$}}\\
\end{array}
\right]
\in \mathbb{R}^{(N^v+N^r) \times (N^v+N^r)}.
\end{equation}

\noindent $\mathcal{K}(\mathbf{S}^{vv}, k_{tr})$ denotes a $k_{tr}$-nearest neighbor chosen operation, where $\mathcal{K}(\mathbf{S}^{vv}, k_{tr})_{ij}=\mathbf{S}^{vv}_{ij}$ if $\mathbf{f}^v_j \in \mathcal{N}(\mathbf{f}^v_i, k_{tr})$ and $\mathbf{S}^{vv}_{ij} > 0$. $\mathcal{N}(\mathbf{f}^v_i, k_{tr})$ is the $k_{tr}$-th nearest neighbor set of $\mathbf{f}^v_i$. $\overline{(\cdot)}$ denotes the row normalize. 
It is noteworthy that the four matrices are normalized separately, thus alleviating the modality misalignment. 
We utilize the global affinity matrix $\mathbf{G}_{tr}$ to propagate the original features:

\vspace{-2mm}
\begin{equation}\label{eq:FP}
\begin{aligned}
    & [\tilde{\mathbf{f}}^v_1, \cdots, \tilde{\mathbf{f}}^v_{N^v}, \tilde{\mathbf{f}}^r_1, \cdots, \tilde{\mathbf{f}}^r_{N^r}]^{\top} = \\ & \mathbf{G}_{tr} 
    [\mathbf{f}^v_1, \cdots, \mathbf{f}^v_{N^v}, \mathbf{f}^r_1, \cdots, \mathbf{f}^r_{N^r}]^{\top}.
\end{aligned}
\end{equation}

\noindent The propagated features $\{\tilde{\mathbf{f}}^v_i\}_{i=1}^{N^v}$ and $\{\tilde{\mathbf{f}}^r_i\}_{i=1}^{N^r}$ are then utilized cross-modality association according to Eq.\ref{eq:OTLA}.

The propagation process can also be integrated into the testing stage. The affinity matrix $\mathbf{G}_{te}$ can be constructed based on the query set $\{\mathbf{q}_i\}_{i=1}^{N^q}$ with $N^q$ queries, the gallery set $\{\mathbf{g}_i\}_{i=1}^{N^g}$ with $N^g$ galleries and the parameter $k_{te}$ for k-nearest neighbor chosen. The propagated features $\{\tilde{\mathbf{q}}_i\}_{i=1}^{N^q}$ and $\{\tilde{\mathbf{g}}_i\}_{i=1}^{N^g}$ are utilized in the final retrieval.

\subsection{optimization}

The overall framework is trained by minimizing:

\begin{equation}\label{eq:total}
    \mathcal{L} = \mathcal{L}_{dagl} + \mathcal{L}_{fgsal} + \lambda \mathcal{L}_{gpcr},
\end{equation}

\noindent where $\lambda$ is a hyper-parameter to balance the loss terms. The CMFP module works in both the dual association stage and the testing stage.
The training process in one epoch is illustrated in Alg.\ref{alg:training}.

\begin{algorithm}[t] 
  \caption{Training process in one epoch.}  
  \label{alg:training}  
    {\textbf{Input:} network $f_{\theta}$; iteration number from $T_0$ to $T_t$ in one epoch; batch size $B$;\\}
    1: \textbf{Extract} visible features $\left\{\mathbf{f}^v_i | i=1, 2, \cdots, N^v\right\}$ and infrared features $\left\{\mathbf{f}^r_i | i=1, 2, \cdots, N^r\right\}$;\\
    2: \textbf{Cluster} features by DBSCAN and obtain the intra-modality pseudo-labels $\left\{\tilde{y}^v_i | i=1, 2, \cdots, N^v\right\}$ and $\left\{\tilde{y}^r_i | i=1, 2, \cdots, N^r\right\}$;\\
    3: \textbf{Initialize} the global-level and part-level memory banks by cluster centroids; \\
    4: \textbf{Do CMFP} as Eq.\ref{eq:G_tr} and Eq.\ref{eq:FP} obtain the propagated features $\{\tilde{\mathbf{f}}^v_i | i=1, 2, \cdots, N^v\}$ and $\{\tilde{\mathbf{f}}^r_i | i=1, 2, \cdots, N^r\}$;\\
    5: Derive modality-unified pseudo-labels $\left\{\hat{y}^v_i | i=1, 2, \cdots, N^v\right\}$ and $\left\{\hat{y}^r_i | i=1, 2, \cdots, N^r\right\}$ by \textbf{dual associations} as Eq.\ref{eq:OTLA};\\
    6: 
    \For{$t = T_0$ to $T_t$}{
    1: Extract feature maps and global features for each instance in a mini-batch;\\
    2: Obtain semantic-aligned features from feature maps $\mathbf{F}^v$ and $\mathbf{F}^r$ of cross-modality positive pairs as Eq.\ref{eq:query}, Eq.\ref{eq:PA1} and Eq.\ref{eq:PA2} ;\\
    3: Compute part contrastive loss $\mathcal{L}_{fgsal}$ as Eq.\ref{eq:total_pacl};\\
    4: Search reliable positive samples for global and part features and compute $\mathcal{L}_{gpcr}$ as Eq.\ref{eq:total_gpcr};\\
    5: Compute global contrastive loss $\mathcal{L}_{dagl}$ as Eq.\ref{eq:total_base};\\
    6: Compute total loss as Eq.\ref{eq:total};\\
    }
    7: \textbf{Update} network $f_{\theta}$ and update prototypes in memory banks as Eq.\ref{eq:update_momentum}.
\end{algorithm}

\section{Experiments}

\begin{table*}[t]
    \centering	
    \renewcommand\arraystretch{1.1}
    \caption{Comparison with the state-of-the-art methods on SYSU-MM01 and RegDB. ``GUR$^\ast$" denotes GUR without camera labels. 
    Since our method does not require any camera labels, for fair comparison we do not report the results of GUR with camera labels.
    The best results are in \textbf{bold}, and the second best results are \underline{underlined}.}
    \begin{adjustbox}{max width=\textwidth}
    \footnotesize
    \begin{tabular}{c|c|ccc|ccc|ccc|ccc}
    \hline
        \multicolumn{1}{c|}{
        \multirow{3}{*}{Method}} &  \multicolumn{1}{c|}{
        \multirow{3}{*}{Venue}} &
    \multicolumn{6}{c|}{SYSU-MM01 (Single-shot)} & 
        \multicolumn{6}{c}{RegDB}\\
        \cline{3-14}
        & & \multicolumn{3}{c|}{All-search} & \multicolumn{3}{c|}{Indoor-search} &
        \multicolumn{3}{c|}{Visible-to-Infrared} & \multicolumn{3}{c}{Visible-to-Infrared}\\ 
        \cline{3-14}
    & & R1 & mAP & mINP & R1 & mAP & mINP & R1 & mAP & mINP &R1 & mAP & mINP\\
        \hline
        \multicolumn{12}{l}{\textit{Supervised VI-ReID methods}} \\ \hline
        Zero-Pad\cite{SYSU-MM01} & ICCV-17 & 14.80 & 15.95 & - & 20.58 & 26.92 & - & 17.75 & 18.90 & - & 16.63 & 17.82 & -\\
        AlignGAN\cite{AlignGAN} & ICCV-19 & 42.4 & 40.7 & - & 45.9 & 54.3 & - & 57.9 & 53.6 & - & 56.3 & 53.4 & - \\
        cm-SSFT \cite{cm-SSFT} & CVPR-20 & 47.7 & 54.1 & - & - & - & - & 72.3 & 72.9 & - & 71.0 & 71.7 & - \\
        DDAG \cite{DDAG} & ECCV-21 & 54.75 & 53.02 & 39.62 & 61.02 & 67.98 & 62.61 & 69.34 & 63.46 & 49.24 & 68.06 & 61.80 & 48.62 \\
        AGW \cite{agw} & TPAMI-21 & 47.50 & 47.65 & 35.30 & 54.17 & 62.97 & 59.23 & 70.05 & 66.37 & 50.19 & 70.49 & 65.90 & 51.24\\
        VCD+VML \cite{VCD+VML} & CVPR-21 & 60.02 & 58.80 & - & 66.05 & 72.98 & - & 73.2 & 71.6 & - & 71.8 & 70.1 & - \\
       CA \cite{CA} & ICCV-21 & 69.88 & 66.89 & 53.61 & 76.26 & 80.37 & 76.79 & 85.03 & 79.14 & 65.33 & 84.75 & 77.82 & 61.56 \\
        MPANet \cite{MPANet} & CVPR-21 & 70.58 & 68.24 & - & 76.74 & 80.95 & - & 82.8 & 80.7 & - & 83.7 & 80.9 & -\\
        LUPI \cite{LUPI} & ECCV-22 & 71.1 &67.6 & - & 82.4 & 82.7 & - & 88.0 & 82.7 & - & 86.8 & 81.3 &-\\
        CMT \cite{CMT} & ECCV-22 & 71.88 & 68.57 & - & 76.9 & 79.91 & - & 96.17 & 87.3 & - & 91.97 & 84.46 & - \\
        CIFT \cite{CIFT} & ECCV-22 & 74.08 & 74.79 & - & 81.82 & 85.61 & - & 91.96 & 92.00 & - & 90.30 & 90.78 & - \\
        DEEN\cite{DEEN} & CVPR-23 & 74.7 & 71.8 & - & 80.3 & 83.3 & - & 91.1 & 85.1 & - & 89.5 & 83.4 & - \\
        SEFEL \cite{SEFEL} & CVPR-23 & 77.12 & 72.33 & - & 82.07 & 82.95 & - & 91.07 & 85.23 & - & 92.18 & 86.59 & - \\
        PartMix \cite{PartMix} & CVPR-23 & 77.78 & 74.62 & - & 81.52 & 84.38 & - & 84.93 & 82.52 & - & 85.66 & 82.27 & - \\
        SAAI \cite{SAAI} & ICCV-23 & 75.90 & 77.03 & - & 83.20 & 88.01 & - & 91.07 & 91.45 & -& 92.09 & 92.01 &- \\
        CAL \cite{CAL} & ICCV-23 & 74.66 & 71.73 & - & 79.69 & 83.68 & - & 94.51 & 88.67 & - & 93.64 & 87.61 & - \\
        IDKL \cite{IDKL} & CVPR-24 & 81.42 & 79.85 & - & 87.14 & 89.37 & - & 94.72 & 90.19 & - & 94.22 & 90.43 & -\\
        \hline
        \multicolumn{12}{l}{\textit{Semi-supervised VI-ReID methods}} \\ \hline
        OTLA \cite{OTLA} & ECCV-22 
        & 48.2 & 43.9 & - & 47.4 & 56.8 &-
        & 49.9 & 41.8 & - & 49.6 & 42.8 &-  \\
        DPIS \cite{DPIS} & ICCV-23 & 58.4 & 55.6 & - & 63.0 & 70.0 & - & 62.3 & 53.2 & - & 61.5 & 52.7 & - \\
        \hline
        \multicolumn{12}{l}
        {\textit{Unsupervised VI-ReID methods}} \\ \hline
        H2H \cite{H2H} & TIP-21 & 25.49 & 25.16 & - & - & - & - & 13.91 & 12.72 & - & 14.11 & 12.29 & - \\
        H2H(AGW) \cite{H2H} & TIP-21 & 30.15 & 29.40 & - & - & - & - & 23.81 & 18.87 & - & - & - & - \\
        H2H(AGW) w/ CMRR \cite{H2H} & TIP-21 & 45.47 & 47.99 & - & - & - & - & 35.18 & 36.46 & - & - & - & - \\
        OTLA \cite{OTLA} & ECCV-22 & 29.9 & 27.1 & - & 29.8 & 38.8 & - & 32.9 & 29.7 & - & 32.1 & 28.6 & - \\
        ADCA \cite{ADCA} & MM-22 & 45.51 & 42.73 & 28.29 & 50.60 & 59.11 & 55.17 & 67.20 & 64.05 & 52.67 & 68.48 & 63.81 & 49.62 \\
        ADCA(AGW) \cite{ADCA} & MM-22 & 50.90 &  45.70 & 29.12 & 51.39 & 59.82 & 56.08 & 66.62 & 63.47 & - & 67.29 & 62.98 & -\\
        TAA \cite{TAA} & TIP-23 & 48.77 & 42.33 & 25.37 & 50.12 & 56.02 & 49.96 & 62.23 & 56.00 & 41.51 & 63.79 & 56.53 & 38.99 \\
        CHCR (AGW) \cite{CHCR} & T-CSVT-23 & 47.72 & 45.34 & - & - & - & - & 68.18 & 63.75 & - & 69.96 & 65.87 & - \\ 
        DOTLA (AGW) \cite{DOTLA} & MM-23 & 50.36 & 47.36 & 32.40 & 53.47 & 61.73 & 57.35 & 85.63 & 76.71 & 61.58 & 82.91 & 74.97 & 58.60 \\
        MBCCM \cite{MBCCM} & MM-23 & 53.14 & 48.16 & 32.41 & 55.21 & 61.98 & 57.23 & 83.79 & 77.87 & 65.04 & 82.82 & 76.74 & 61.73 \\
        CCLNet (CLIP) \cite{CCLNet} & MM-23 & 
        54.03 & 50.19 & - & 56.68 & 65.12 & - &69.94 & 65.53 & - & 70.17 & 66.66 & -\\
        PGM(AGW) \cite{PGM} & CVPR-23 & 57.27 & 51.78 & 34.96 & 56.23 & 62.74 & 58.13 & 69.48 & 65.41 & - & 69.85 & 65.17 & - \\
    GUR$^\ast$ \cite{GUR} & ICCV-23 &
    60.95 & 56.99 & 41.85 & 64.22 & 69.49 & 64.81 & 73.91 & 70.23 & 58.88 & 75.00 & 69.94 & 56.21 \\
    MIMR \cite{MIMR} & KBS-24 & 46.56 & 45.88 & - & 52.26 & 60.93 & - & 68.76 & 64.33 & - & 68.76 & 63.83 & -\\
    \textcolor{black}{IMSL} \cite{ISML-TCSVT24} & \textcolor{black}{T-CSVT-24} & 
    \textcolor{black}{57.96}	&
    \textcolor{black}{53.93}	& 
    \textcolor{black}{-}	&
    \textcolor{black}{58.30}	&
    \textcolor{black}{64.31}	&
    \textcolor{black}{-	}&
    \textcolor{black}{70.08}	&
    \textcolor{black}{66.30}	&
    \textcolor{black}{-}	&
    \textcolor{black}{70.67}	&
    \textcolor{black}{66.35}	&
    \textcolor{black}{- }\\
    \textcolor{black}{BCGM} \cite{BCGM-MM24} & \textcolor{black}{MM-24} &  \textcolor{black}{61.7} & \textcolor{black}{56.1} & \textcolor{black}{38.7} & \textcolor{black}{60.9} & \textcolor{black}{66.5} & \textcolor{black}{62.3} & \textcolor{black}{86.8} & \textcolor{black}{81.7} & \textcolor{black}{68.6} & \textcolor{black}{86.7} & \textcolor{black}{82.3} & \textcolor{black}{\underline{71.1}} \\
    \textcolor{black}{MMM} \cite{MMM} & \textcolor{black}{ECCV-24} & \textcolor{black}{61.60} & \textcolor{black}{57.90} & \textcolor{black}{--} & 
    \textcolor{black}{64.40} & \textcolor{black}{70.40} & \textcolor{black}{--} & 
    \textcolor{black}{89.70} & \textcolor{black}{80.50} & \textcolor{black}{--} & 
    \textcolor{black}{85.80} & \textcolor{black}{77.00} & \textcolor{black}{--} \\
    \textcolor{black}{PCLHD} \cite{PCMIP} & \textcolor{black}{NIPS-24} & \textcolor{black}{64.40} & \textcolor{black}{58.70} & \textcolor{black}{--} & \textcolor{black}{\underline{69.50}} & \textcolor{black}{\underline{74.40}} & \textcolor{black}{--} & 
    \textcolor{black}{84.30} & \textcolor{black}{80.70} & 
    \textcolor{black}{--} &
    \textcolor{black}{82.70} & \textcolor{black}{78.40} & \textcolor{black}{--} \\
    \textcolor{black}{PCAL} \cite{PCAL-TIFS25} & \textcolor{black}{TIFS-25} & 
    \textcolor{black}{57.94}	&
    \textcolor{black}{52.85}	&
    \textcolor{black}{36.90}	&
    \textcolor{black}{60.07}	&
    \textcolor{black}{66.73}	&
    \textcolor{black}{62.09}	&
    \textcolor{black}{86.43}	&
    \textcolor{black}{82.51}	&
    \textcolor{black}{\underline{72.33}}	&
    \textcolor{black}{86.21}	&
    \textcolor{black}{81.23}	&
    \textcolor{black}{68.71} \\
    \hline
    SALCR (ours) & - & \underline{64.44} & \underline{60.44} & \underline{45.19} & 67.17 & 72.88 & \underline{68.73} & \underline{90.58} & \underline{83.87} & 70.76 & \underline{88.69} & \underline{82.66} & 66.89 \\
    SALCR w/ CMFP(te) (ours) & - & \textbf{78.29} & \textbf{74.08} & \textbf{60.68} & \textbf{79.64} & \textbf{82.75} & \textbf{79.65} & \textbf{93.01} & \textbf{93.16} & \textbf{91.04} & \textbf{94.08} & \textbf{93.46} & \textbf{90.73} \\
    \hline	
    \end{tabular}
    \end{adjustbox}
    \vspace{0mm}
\label{tab:sota_comparison}
\end{table*}

\subsection{Datasets and Evaluation Protocol}

\textbf{Datasets.} We evaluate the proposed method on two public visible-infrared ReID datasets, namely SYSU-MM01 \cite{SYSU-MM01} and RegDB \cite{RegDB}. The trainset of SYSU-MM01 contains 395 identities, with 22258 visible images and 11909 infrared images captured from four visible cameras and two infrared cameras. The testing set of SYSU-MM01 contains 95 identities, with 3803 infrared query
images and 301 visible gallery images. We adopt All-Search and Indoor-Search modes under single-shot and multi-shot settings. 
In All-Search mode, the gallery set contains all visible images from both indoor and outdoor cameras, while in
Indoor-Search mode, the gallery set only contains images captured from the indoor environment.
We strictly follow existing methods with 10 randomly gallery set selections \cite{CA}, and report the average performance.

RegDB is a smaller-scale dataset captured from a pair of aligned visible and infrared cameras \cite{RegDB}. It contains 412 identities, where each identity holds 10 visible images and 10 infrared images. Following \cite{CA}, we randomly select 206 identities for training and the remaining identities for testing. 
This procedure is repeated 10 times and the average performance is reported. 
The model is evaluated under two modes: Visible-to-Infrared and Infrared-to-Visible.

\textbf{Evaluation Metrics.} All experiments follow the common evaluation protocols used in USL-VI-ReID \cite{PGM, MBCCM, GUR}. The evaluation metrics include Cumulative Matching Characteristic (CMC), Mean Average Precision (mAP), and Mean Inverse Negative Penalty (mINP \cite{agw}).

\begin{table}
    \centering	
    \renewcommand\arraystretch{1.1}
    \caption{Comparison with the state-of-the-art methods under the Multi-shot setting on SYSU-MM01.
    The best results are in \textbf{bold}, and the second best results are \underline{underlined}.}
    \begin{adjustbox}{max width=0.48\textwidth}
    \footnotesize
    \begin{tabular}{c!{\vrule width0.5pt}c!{\vrule width0.5pt}cc!{\vrule width0.5pt}cc}
    \Xhline{0.5pt}
        \multicolumn{1}{c!{\vrule width0.5pt}}{
        \multirow{3}{*}{Method}} &
        \multicolumn{1}{c!{\vrule width0.5pt}}{
        \multirow{3}{*}{Venue}} &
        \multicolumn{4}{c}{SYSU-MM01 (Multi-shot)} \\
        \Xcline{3-6}{0.5pt}
        & & \multicolumn{2}{c!{\vrule width0.5pt}}{All-search} & \multicolumn{2}{c}{Indoor-search}\\ 
        \Xcline{3-6}{0.5pt}
    & & R1 & mAP & R1 & mAP\\
        \Xhline{0.5pt}
        \multicolumn{4}{l}{\textit{Supervised VI-ReID methods}} \\ \Xhline{0.5pt}
        Zero-Pad \cite{SYSU-MM01} & ICCV-17 & 19.13 & 10.89 & 24.43 & 18.86 \\
        cmGAN \cite{cmGAN}  & IJCAI-18 & 31.49 & 22.27 & 37.00 & 32.76\\
        AlignGAN \cite{AlignGAN} & ICCV-19  & 51.50 & 33.90 & 57.10 & 45.30 \\
        cm-SSFT \cite{cm-SSFT} & CVPR-20 & 63.40 & 62.00 & 73.00 & 72.40 \\
        MPANet \cite{MPANet} & CVPR-21 & 75.58 & 62.91 & 84.22 & 75.11 \\
        CMT \cite{CMT} & ECCV-22 & 80.23 & 63.13 & 84.87 & 74.11 \\
        CIFT \cite{CIFT} & ECCV-22 & 79.74 & 75.56 & 88.32 & 86.42 \\
        PartMix \cite{PartMix} & CVPR-23 & 80.54 & 69.84 & 87.99 & 79.95\\
        SAAI \cite{SAAI} & ICCV-23 & 82.86 & 82.39 & 90.73 & 91.30\\
        CAL \cite{CAL} & ICCV-23 & 77.05 & 64.86 & 86.97 & 78.51 \\
        IDKL \cite{IDKL} & CVPR-24 & 84.34 & 78.22 & 94.30 & 88.75\\
        \Xhline{0.5pt}
        \multicolumn{4}{l}
        {\textit{Unsupervised VI-ReID methods}} \\ \Xhline{0.5pt}
        H2H \cite{H2H} & TIP-21 & 30.31	& 19.12 & - & - \\
        DFC \cite{DFC} & IPM-23 & 44.12 & 28.36 & - & -\\
        MBCCM \cite{MBCCM} & MM-23 &57.73 & 39.78 & 62.87 & 52.80 \\
        CHCR \cite{CHCR} & T-CSVT-23 & 50.12 & 42.17 & - & - \\
        MULT \cite{he2024exploring} & IJCV-24 & 71.35 & 52.18 & 76.99 & 64.03 \\
        \Xhline{0.5pt}
    SALCR (ours) & - & \underline{72.09} & \underline{53.79} & \underline{77.49} & \underline{65.36}\\
    SALCR w/ CMFP(te) (ours) & - & \textbf{80.95} & \textbf{74.60} & \textbf{86.72} & \textbf{83.52}\\
    \Xhline{0.5pt}
    \end{tabular}
    \end{adjustbox}
    \vspace{-3mm}
\label{tab:multi-shot-comparison}
\end{table}

\begin{table*}[!htbp]
    \centering	
    \renewcommand\arraystretch{1.1}
    \caption{Ablation study on individual components of our method on SYSU-MM01.}
    \begin{adjustbox}{max width=\textwidth}
    \footnotesize
    \begin{tabular}{c!{\vrule width0.5pt}ccccc!{\vrule width0.5pt}ccc!{\vrule width0.5pt}ccc!{\vrule width0.5pt}ccc!{\vrule width0.5pt}ccc}
    \Xhline{0.5pt} 
\multicolumn{1}{c!{\vrule width0.5pt}}{
\multirow{3}{*}{Idx}} &
\multicolumn{5}{c!{\vrule width0.5pt}}{
\multirow{2}{*}{Components}} &
\multicolumn{6}{c!{\vrule width0.5pt}}{SYSU-MM01 (single-shot)} &
\multicolumn{6}{c}{RegDB}\\
\Xcline{7-18}{0.5pt}
& & & & & & \multicolumn{3}{c!{\vrule width0.5pt}}{All-search} & \multicolumn{3}{c|}{Indoor-search} &
\multicolumn{3}{c!{\vrule width0.5pt}}{Visible-to-Infrared} &
\multicolumn{3}{c}{Infrared-to-Visible} \\ 
\Xcline{2-18}{0.5pt}
& B & DAGL & FGSAL & GPCR & CMFP(tr) & R1 & mAP & mINP & R1 & mAP & mINP & R1 & mAP & mINP & R1 & mAP & mINP \\
\Xhline{0.5pt}
1 & \checkmark & & & & & 40.83 & 39.65 & 26.08 & 44.81 & 53.72 & 49.49 & 
50.28 & 49.34 & 35.88 & 50.17 & 47.32 & 33.85 \\
2 & \checkmark& \checkmark & & & & 58.89 & 53.87 & 37.50 & 59.87 & 66.17 & 61.55 
& 86.80 & 79.75 & 65.49 & 85.28 & 78.52 & 61.87 \\
3 & \checkmark& \checkmark & \checkmark & & & 61.81 & 57.21 & 41.38 & 64.07 & 69.42 & 64.64
& 89.37 & 82.94 & 69.16 & 87.01 & 81.22 & 65.60\\
4 & \checkmark& \checkmark & & \checkmark & & 59.83 & 55.61 & 39.86 & 59.62 & 66.84 & 62.58
& 87.62 & 81.46 & 67.92 & 86.94 & 80.61 & 64.46\\
5 & \checkmark& \checkmark & \checkmark & \checkmark & & 63.25 & 58.64 & 42.64 & 66.20 & 71.30 & 66.76 
& 90.05 & 83.23 & 70.36 & 87.86 & 82.07 & 66.34\\
\Xhline{0.5pt}
6 & \checkmark& \checkmark & & & \checkmark & 60.12 & 55.38 & 39.57 & 60.98 & 66.98 & 62.52
& 87.23 & 80.55 & 66.78 & 86.67 & 79.44 & 62.58\\
7 & \checkmark& \checkmark & \checkmark & & \checkmark & 62.07 & 58.48 & 43.40 & 64.29 & 71.30 & 67.47 
& 89.71 & 83.22 & 69.91 & 87.57 & 82.04 & 65.56\\
8 & \checkmark& \checkmark & & \checkmark & \checkmark & 60.38 & 56.36 & 40.81 & 59.70 & 67.24 & 63.01 
& 89.09 & 82.44 & 69.05 & 88.25 & 81.31 & 65.95
\\
9 & \checkmark& \checkmark & \checkmark & \checkmark & \checkmark &  \textbf{64.44} & \textbf{60.44} & \textbf{45.19} & \textbf{67.17} & \textbf{72.88} & \textbf{68.73} 
& \textbf{90.58} & \textbf{83.87} & \textbf{70.76} &
\textbf{88.69} & \textbf{82.66} & \textbf{66.89}\\
\Xhline{0.5pt}
\end{tabular}
\end{adjustbox}
\label{tab:ablation}
\end{table*}

\subsection{Implementation Details}

The proposed method is implemented using PyTorch. The two-stream ResNet \cite{resnet} is employed as the backbone, which is pretrained on ImageNet \cite{ImageNet}. The model is trained for a total of 80 epochs.
The Adam Optimizer is adopted with a weight decay of 5e-4.
The initial learning rate is 3.5e-4 and decays 10 times at the 20-th and 60-th epochs.
In the first 40 epochs, the model is trained under the DCL framework \cite{ADCA}, and our SALCR framework is executed in another 40 epochs. 
All the images are resized into $288 \times 144$.
Random horizontal flipping, random erasing, random cropping, random color jitter, and random channel augmentation \cite{CA} are utilized as data augmentations following \cite{PGM}. Additionally, the LTG \cite{LTG} is adopted in the DCL training stage to promote learning color-irrelevant features. 
During DCL training, We sample 12 identities from both modalities and 12 images for each identity in a mini-batch.
While training with our SALCR, we sample 8 pseudo-labels with 4 visible images and 4 infrared images for each label in a mini-batch. 
Following \cite{ClusterContrast, PGM, ADCA}, the momentum factor $\mu$ and the temperature factor $\tau$ are set to 0.1 and 0.05, the maximum distance for DBSCAN is set to 0.6 for SYSU-MM01 and 0.3 for RegDB.
The hyper-parameter $\lambda_{ot}$ for the dual association is set to 5.
The number of parts $N_p$ in FGSAL is set to 3 and $k$ for nearest neighbor search in GPCR is set to 30.
The $k_{tr}$ and $k_{te}$ in CMFP during training and testing are both 30 for SYSU-MM01 and 8 for RegDB.
The trade-off parameter $\lambda$ for the loss terms is set to 0.5.

\subsection{Comparison with State-of-the-Art Methods}

We compare the proposed method with state-of-the-art methods on SYSU-MM01 and RegDB datasets under three relevant settings, $i.e.$, supervised VI-ReID, semi-supervised VI-ReID, and unsupervised VI-ReID. The results are shown in Tab.\ref{tab:sota_comparison} and Tab.\ref{tab:multi-shot-comparison}. ``CMFP(te)'' in Tab.\ref{tab:sota_comparison} and Tab.\ref{tab:multi-shot-comparison} corresponds to the performance of conducting CMFP as a re-ranking stage after feature extraction during testing.

\emph{(a) Comparison with Unsupervised Methods:} 
As reported in Tab.\ref{tab:sota_comparison}, our method outperforms the state-of-the-art GUR$^\ast$ \cite{GUR} by a margin of 3.45\% mAP and 3.49\% Rank1 on SYSU-MM01 (All-Search), and 13.64\% mAP and 16.67\% Rank1 on RegDB (Visible-to-Infrared).
Additionally, with the incorporation of CMFP during testing, our method achieves the surprising performance of 74.08\% mAP and 78.29\% Rank1, exhibiting superior improvement compared to existing methods. 
As reported in Tab.\ref{tab:multi-shot-comparison}, our methods surpass CHCR \cite{CHCR} by a considerable margin of 11.62\% mAP and 21.97\% Rank1 on SYSU-MM01 (All-Search) under the multi-shot setting.
Existing methods \cite{PGM, MBCCM, GUR} mainly focus on learning modality-unified global representations, while neglecting complicated part-level cross-modality interactions. 
Our method explores part-level semantic-aligned cross-modality pairs and builds up optimization objectives for learning fine-grained modality-shared information.
Such a learning paradigm fully exploits the complementary pseudo-label spaces and promotes discovering more abundant intra-cluster details.


\emph{(b) Comparison with Semi-Supervised Methods:} The proposed method outperforms the state-of-the-art DPIS \cite{DPIS} by 4.84\% mAP and 6.04\% Rank1.
In the semi-supervised setting, visible ground-truth labels are accessible for training. 
The results demonstrate the potential of purely unsupervised methods, which do not require any annotation and are more data-friendly in real scenarios.

\emph{(c) Comparison with Supervised Methods:} We additionally provide results of 17 well-known supervised methods for reference in Tab.\ref{tab:sota_comparison}. Our method surpasses several supervised methods including Zero-Pad \cite{SYSU-MM01}, AlignGAN \cite{AlignGAN}, cm-SSFT \cite{cm-SSFT}, DDAG \cite{DDAG}, AGW \cite{agw} and VCD+VML \cite{VCD+VML} on SYSU-MM01 (All-Search). For the easier dataset RegDB, our method achieves a surprising performance that approaches some of the latest methods ($e.g.$, DEEN \cite{DEEN} and PartMix \cite{PartMix}). However, there is still a significant margin compared to SOTAs.

\begin{table}[!htbp]
\vspace{-4mm}
    \centering	
    \renewcommand\arraystretch{1.1}
    \caption{Comparison with uni-directional association on SYSU-MM01.}
    \begin{adjustbox}{max width=0.48\textwidth}
    \footnotesize
    \begin{tabular}{c|ccc|ccc}
    \hline
\multicolumn{1}{c|}{
\multirow{2}{*}{Method}}
& \multicolumn{3}{c|}{All-Search} & \multicolumn{3}{c}{Indoor-Search} \\ 
\cline{2-7}
& R1 & mAP & mINP & R1 & mAP & mINP \\
\hline
Visible branch & 53.17 & 47.61 & 30.68 & 54.36 & 61.64 & 56.92 \\
Infrared branch & 55.67 & 51.90 & 36.47 & 57.61 & 64.81 & 60.31 \\
Dual & \textbf{58.89} & \textbf{53.87} & \textbf{37.50} & \textbf{59.87} & \textbf{66.17} & \textbf{61.55} \\
\hline
\end{tabular}
\end{adjustbox}
\label{tab:uni-dual}
\vspace{-6mm}
\end{table}

\begin{table}[!htbp]
\vspace{-4mm}
\centering	
\renewcommand\arraystretch{1.1}
\caption{Comparison with other re-ranking methods on SYSU-MM01.}
\begin{adjustbox}{max width=0.48\textwidth}
\footnotesize
\begin{tabular}{c!{\vrule width0.5pt}c!{\vrule width0.5pt}cc!{\vrule width0.5pt}cc!{\vrule width0.5pt}c}
\Xhline{0.5pt}
\multicolumn{1}{c!{\vrule width0.5pt}}{
\multirow{2}{*}{Method}} &
\multicolumn{1}{c!{\vrule width0.5pt}}{
\multirow{3}{*}{Venue}} &\multicolumn{2}{c!{\vrule width0.5pt}}{All-search} & \multicolumn{2}{c!{\vrule width0.5pt}}{Indoor-search} & \multicolumn{1}{c}{\multirow{2}{*}{Time}} \\ 
\Xcline{3-6}{0.5pt}
& & R1 & mAP & R1 & mAP & \\
\Xhline{0.5pt}
None & - & 64.44 & 60.44 & 67.17 & 72.88 &-\\
CMRR (k=15) & TIP-21 & \underline{72.10} & \underline{69.14} & \underline{71.85} & 73.26 & 12.260s \\
CMRR (k=25) & TIP-21 & 70.83 & 67.89 & 70.31 & \underline{75.87} & 19.247s\\
AIM & ICCV-23 & 64.80 & 64.23 & 67.51 & 74.40 & \textbf{0.053s} \\
CMFP(te) (Ours) & - & \textbf{78.29} & \textbf{74.08} & \textbf{79.64} & \textbf{82.75} & \underline{0.241s} \\
\Xhline{0.5pt}
\end{tabular}
\end{adjustbox}
\label{tab:reranking-comparison-sysu}
\vspace{-4mm}
\end{table}

\begin{table}[!htbp]
\vspace{-4mm}
\centering	
\renewcommand\arraystretch{1.1}
\caption{Comparison with other re-ranking methods on RegDB.}
\begin{adjustbox}{max width=0.48\textwidth}
\footnotesize
\begin{tabular}{c!{\vrule width0.5pt}c!{\vrule width0.5pt}cc!{\vrule width0.5pt}cc!{\vrule width0.5pt}c}
\Xhline{0.5pt} 
\multicolumn{1}{c!{\vrule width0.5pt}}{
\multirow{2}{*}{Method}} &
\multicolumn{1}{c!{\vrule width0.5pt}}{
\multirow{3}{*}{Venue}} & \multicolumn{2}{c!{\vrule width0.5pt}}{V-to-I} & \multicolumn{2}{c!{\vrule width0.5pt}}{I-to-V} & \multicolumn{1}{c}{\multirow{2}{*}{Time}} \\ 
\Xcline{3-6}{0.5pt}
& & R1 & mAP & R1 & mAP & \\
\Xhline{0.5pt}
None & - & 90.58 & 83.87 & 88.69 & 82.66 &-\\
CMRR (k=8) & TIP-21 & \textbf{95.39} & \textbf{95.72} & \textbf{95.73} & \textbf{95.62} & 8.73s \\
CMRR (k=15) & TIP-21 & 91.07 & 92.35 & 90.00 & 91.46 & 12.016s\\
AIM & ICCV-23 & 89.71 & 89.50 & 89.85 & 89.49 & \textbf{0.053s} \\
CMFP(te) (Ours) & - & \underline{93.01} & \underline{93.16} & \underline{94.08} & \underline{93.46} & \underline{0.488s} \\
\Xhline{0.5pt}
\end{tabular}
\end{adjustbox}
\label{tab:reranking-comparison-regdb}
\vspace{-5mm}
\end{table}

\subsection{Ablation Study}

To demonstrate the effectiveness of each component of our method, we conduct ablation experiments on SYSU-MM01 and RegDB. The results are shown in Tab.\ref{tab:ablation}. ``CMFP(tr)" indicates we apply CMFP for features extracted from the trainset before the cross-modality association.
``B" denotes our intra-modality DCL baseline Sec.\ref{baseline}, which achieves 39.65\% mAP on SYSU-MM01.

\emph{(a) The Effectiveness of DAGL:} 
The DAGL module improves performance by 14.13\% mAP on SYSU-MM01 (All-Search) compared to the baseline, indicating the superiority of integrating such an OTLA-based dual association approach into the memory-based framework. 
OTLA utilizes detailed instance-cluster relationships for associations, thus attaining better performance compared to several existing methods ($i.e.$, PGM \cite{PGM} and MBCCM \cite{MBCCM}) that only consider cluster-level relationships.
We further compare our DAGL with uni-directional OTLA designs, as shown in Tab.\ref{tab:uni-dual}.
``Visible branch'' denotes only pseudo-labels assigned from the visible modality $\{\hat{y^r_i}\}_{i=1}^{N^r}$ are involved in the cross-modality interaction learning, and vice versa.
The results demonstrate the effectiveness of the dual association.
Compared to the uni-directional method, our dual association incorporates more implicit potential associations into training, which could effectively multigate the side-effects raised from some insufficient label associations.





\emph{(b) The Effectiveness of FGSAL:} 
Our FGSAL achieves an improvement of 3.34\% mAP without CMFP(tr) and 3.10\% mAP with CMFP(tr) on SYSU-MM01 (All-Search) when directly applied to our DAGL module.   
It further improves performance by 3.03\% mAP without CMFP(tr) and by 4.08\% mAP with CMFP(tr) when collaborating with the GPCR module.
The FGSAL module exploits the complementary nature of label spaces and formulates optimization for modality-related semantic-aligned part features, significantly boosting the performance.
It facilitates complicated cross-modality interactions between part-level features, enabling the model to explore muti-faceted intra-cluster information and richer details about cluster boundaries.



\emph{(c) The Effectiveness of GPCR:}
GPCR further enhances mAP by 1.43\% without CMFP(tr) and by 1.96\% with CMFP(tr) when working together with FGSAL. 
The cluster-level prototypes are not reliable enough due to the inevitable label noise during the early training stage.
Our dynamic GPCR module addresses this issue by adjusting incorrect structure information from noisy labels through the online discovery of positive sets, which can be regarded as a refinement for offline pseudo-labels.
Moreover, our GPCR designs specific strategies for identifying positive samples of global and part features, preventing overfitting to easy instances and enhancing the learning of instance-level relationships.
We further conduct ablation studies for loss terms $\mathcal{L}_g^I$ and $\mathcal{L}_p^I$ ($\mathcal{L}_{v(p)}^I$ \& $\mathcal{L}_{r(p)}^I$) designed for global and part features in GPCR, the results are shown in Tab.\ref{tab:ablation-gpcr}.
The results indicate that the two loss terms both improve model performance and mutually reinforce each other.

\begin{table}[!htbp]
\vspace{-4mm}
\centering	
\renewcommand\arraystretch{1.1}
\caption{Ablation study on loss terms $\mathcal{L}_g^I$ and $\mathcal{L}_p^I$ in GPCR on SYSU-MM01.}
\begin{adjustbox}{max width=0.48\textwidth}
\footnotesize
\begin{tabular}{c|ccc|ccc}
\hline
\multicolumn{1}{c|}{
\multirow{2}{*}{Method}}
& \multicolumn{3}{c|}{All-Search} & \multicolumn{3}{c}{Indoor-Search} \\ 
\cline{2-7}
& R1 & mAP & mINP & R1 & mAP & mINP \\
\hline
w/o GPCR & 62.07 & 58.48 & 43.40 & 64.29 & 71.30 & 67.47 \\
\hline
+ $\mathcal{L}_g^I$ & 63.67 & 59.71 & 44.88 & 66.39 & 72.34 & 68.35\\
+ $\mathcal{L}_p^I$ & 62.83 & 58.97 & 44.34 & 65.78 & 71.85 & 68.01 \\
+ $\mathcal{L}_{gpcr}$ & \textbf{64.44} & \textbf{60.44} & \textbf{45.19} & \textbf{67.17} & \textbf{72.88} & \textbf{68.73} \\
\hline
\end{tabular}
\end{adjustbox}
\label{tab:ablation-gpcr}
\vspace{-4mm}
\end{table}


\emph{(d) The Effectiveness of CMFP:}
From the comparison between the upper section and lower section of Tab.\ref{tab:ablation}, we observe that blending CMFP(tr) into cross-modality associations significantly gains performance on SYSU-MM01. It achieves an improvement of 1.51\% mAP with our DAGI and 1.80\% mAP when applied to the entire SALCR framework on SYSU-MM01. However, the performance gain of CMFP(tr) on RegDB is limited. 
This can be attributed to fewer variations in poses and illumination in RegDB, enabling the model to learn compact intra-cluster representations. 
Such representations are robust enough for the direct instance-cluster associations. 
To demonstrate the efficacy of CMFP during testing,
we further compare our CMFP with existing re-ranking methods for VI-ReID, including CMRR \cite{H2H} and AIM \cite{SAAI} on SYSU-MM01 and RegDB.
We report the performance of CMRR under two different hyper-parameter settings on each dataset, while the hyper-parameter setting for AIM follows \cite{SAAI}.
The average time of various re-ranking algorithms on SYSU-MM01 (All-Search) and RegDB (Visible-to-Infrared) are also reported.  
The results are shown in Tab.\ref{tab:reranking-comparison-sysu} and Tab.\ref{tab:reranking-comparison-regdb}. 
Our CMFP exhibits impressive performance that outperforms existing re-ranking methods by a large margin on SYSU-MM01. 
While on RegDB, CMFP achieves performance approaching CMRR \cite{H2H} but significantly less time consumption compared to CMRR.
Furthermore, our CMFP involves only one hyper-parameter $k_{te}$ during testing, which is more friendly for hyper-parameter tuning in real applications.
We also integrate CMFP as a post-processing approach into existing USL-VI-ReID frameworks, including CNN-based methods PGM \cite{PGM} and MULT \cite{he2024exploring}, as well as the ViT-based method SDCL \cite{sdcl}.
The results are shown in Tab\ref{tab:different-methods-with-cmfp}.

\begin{table}[!htbp]
\centering	
\renewcommand\arraystretch{1.1}
\caption{The effectiveness of CMFP when applied to different USL-VI-ReID methods on SYSU-MM01. ``SDCL$^{\ast}$'' denotes our reproduced results of SDCL \cite{sdcl}.
}
\begin{adjustbox}{max width=0.48\textwidth}
\footnotesize
\begin{tabular}{c|ccc|ccc}
\hline
\multicolumn{1}{c|}{
\multirow{2}{*}{Method}}
& \multicolumn{3}{c|}{All-Search} & \multicolumn{3}{c}{Indoor-Search} \\ 
\cline{2-7}
& R1 & mAP & mINP & R1 & mAP & mINP \\
\hline
PGM & 56.30 & 51.35 & 35.06 & 57.96 & 64.90 & 60.73 \\
PGM w/ CMFP & \textbf{65.58} & \textbf{60.95} & \textbf{46.27} & \textbf{69.89} & \textbf{74.20} & \textbf{70.84}\\
\hline
MULT & 64.77 & 59.23 & 43.46 & 65.34 & 71.46 & 67.83\\
MULT w/ CMFP & \textbf{75.71} & \textbf{72.16} & \textbf{59.65} & \textbf{77.98} & \textbf{81.06} & \textbf{77.60}\\
\hline
SDCL$^{\ast}$ &65.24& 63.64	& 50.84	& 70.92	& 76.77	& 73.35\\
SDCL$^{\ast}$ w/ CMFP 
& \textbf{78.37}	
& \textbf{76.11}	
& \textbf{66.07}	
& \textbf{82.16}	
& \textbf{85.69}	
& \textbf{83.73} \\
\hline
\end{tabular}
\end{adjustbox}
\label{tab:different-methods-with-cmfp}
\vspace{-2mm}
\end{table}

We further provide a theoretical analysis of the efficiency of our CMFP.
For simplicity, we assume that the testing set contains $N$ visible images and $N$ infrared images, respectively.
Specifically, we analyze the computational cost of the three following steps in CMFP:
(a) Computing pairwise similarities between testing images: $\mathcal{O}(N^2d)$;
(b) Sorting the similarity values for $N$ rows and retaining the top-$K_{te}$ entries for each row: $\mathcal{O}(N^2 \log N) + \mathcal{O}(N^2)$;
(c) The feature propagation process based on pairwise similarities and original features: $\mathcal{O}(N^2 d)$.
$d$ represents the dimensional of image features.
Since $\log N \ll d$,
we can derive the overall computational complexity of our CMFP: $\mathcal{O}(N^2d) + \mathcal{O}(N^2 \log N) + \mathcal{O}(N^2) + \mathcal{O}(N^2 d) \approx \mathcal{O}(N^2 d)
$.
Furthermore, our proposed CMFP framework relies exclusively on matrix multiplications, which inherently support parallel operations and ensure computational efficiency.

\begin{table}[!htbp]
\vspace{-6mm}
    \centering	
    \renewcommand\arraystretch{1.1}
    \caption{Parameter analysis of $N_p$ for FGSAL on SYSU-MM01 dataset.}
    \begin{adjustbox}{max width=0.48\textwidth}
    \footnotesize
    \begin{tabular}{c|ccc|ccc}
    \hline 
\multicolumn{1}{c|}{
\multirow{2}{*}{Value of $N_p$}}
& \multicolumn{3}{c|}{All-Search} & \multicolumn{3}{c}{Indoor-Search} \\ 
\cline{2-7}
& R1 & mAP & mINP & R1 & mAP & mINP \\
\hline
$N_p=2$ & \textbf{64.48} & 59.90 & 44.36 & \textbf{68.41} & \textbf{73.59} & \textbf{69.21}\\
$N_p=3$ & 64.44 & \textbf{60.44} & \textbf{45.19} & 67.17 & 72.88 & 68.73 \\
$N_p=4$ & 62.76 & 59.56 & 44.55 & 65.75 & 71.93 & 67.69 \\
\hline
\end{tabular}
\end{adjustbox}
\label{tab:N-p-analysis}
\vspace{-6mm}
\end{table}

\begin{table}[!htbp]
\vspace{-5mm}
    \centering	
    \renewcommand\arraystretch{1.1}{\caption{Parameter analysis of $N_p$ for FGSAL on RegDB dataset.}}
    \begin{adjustbox}{max width=0.48\textwidth}
    \footnotesize
    \begin{tabular}{c|ccc|ccc}
    \hline 
\multicolumn{1}{c|}{
\multirow{2}{*}{Value of $N_p$}}
& \multicolumn{3}{c|}{Visible-to-Infrared} & \multicolumn{3}{c}{Infrared-to-Visible} \\ 
\cline{2-7}
& R1 & mAP & mINP & R1 & mAP & mINP \\
\hline
$N_p=2$ & 88.98	& 82.39	& 69.59	
& 87.29	& 81.03	& 65.55\\
$N_p=3$ & \textbf{90.58} & 
\textbf{83.87} & 70.76 &
\textbf{88.69} & 82.66 & \textbf{66.89}\\
$N_p=4$ & 89.27 & 83.49 & \textbf{71.39} 
& 88.26 & \textbf{82.70} & 66.87\\
\hline
\end{tabular}
\end{adjustbox}
\label{tab:N-p-analysis-regdb}
\vspace{-6mm}
\end{table}

\begin{table}[!htbp]
\centering	
\renewcommand\arraystretch{1.1}
\caption{Parameter analysis of $k_{tr}$ on SYSU-MM01 dataset.}
\begin{adjustbox}{max width=0.48\textwidth}
\footnotesize
\begin{tabular}{c|ccc|ccc}
\hline 
\multicolumn{1}{c|}{
\multirow{2}{*}{Value of $k_{tr}$}}
& \multicolumn{3}{c|}{All-Search} & \multicolumn{3}{c}{Indoor-Search} \\ 
\cline{2-7}
& R1 & mAP & mINP & R1 & mAP & mINP \\
\hline
$k_{tr}=10$ & 62.87 & 58.62 & 42.94 & 64.90 & 70.47 & 65.95 \\
$k_{tr}=20$ & 63.38 & 59.37 & 44.05 & 65.97 & 72.34 & 68.39 \\
$k_{tr}=30$ & \textbf{64.44} & \textbf{60.44} & \textbf{45.19} & 67.17 & \textbf{72.88} & \textbf{68.73} \\
$k_{tr}=40$ & 63.42 & 59.79 & 44.70 & 66.60 & 72.37 & 68.21 \\
$k_{tr}=50$ & 63.21 & 59.60 & 44.66 & \textbf{67.32} & 72.86 & 68.72 \\
\hline
\end{tabular}
\end{adjustbox}
\label{tab:k-tr-analysis}
\vspace{-4mm}
\end{table}

\begin{table}[!htbp]
\centering	
\renewcommand\arraystretch{1.1}
{\caption{Parameter analysis of $k_{tr}$ on RegDB dataset.}}
\begin{adjustbox}{max width=0.48\textwidth}
\footnotesize
\begin{tabular}{c|ccc|ccc}
\hline 
\multicolumn{1}{c|}{
\multirow{2}{*}{Value of $k_{tr}$}}
& \multicolumn{3}{c|}{Visible-to-Infrared} & \multicolumn{3}{c}{Infrared-to-Visible} \\ 
\cline{2-7}
& R1 & mAP & mINP & R1 & mAP & mINP \\
\hline
$k_{tr}=4$ & 88.93	& 82.81	& 69.43	& \textbf{88.79}	&82.19	&65.82 \\
$k_{tr}=8$ & \textbf{90.58}	& \textbf{83.87}	& 70.76	& 88.69	& 82.66	& 66.89 \\
$k_{tr}=15$ & 90.30	& 83.83	& \textbf{71.30} & 88.06	& \textbf{82.81}	& \textbf{67.13} \\
$k_{tr}=20$ & 90.15	& 83.67	& 70.92	& 87.96	& 82.52	& 66.80 \\
$k_{tr}=25$ & 88.74	& 82.45	& 69.30	& 87.53	& 81.43	& 65.39\\
\hline
\end{tabular}
\end{adjustbox}
\label{tab:k-tr-analysis-regdb}
\vspace{-4mm}
\end{table}

\begin{table}[!htbp]
\centering	
\renewcommand\arraystretch{1.1}
\caption{Parameter analysis of $k_{te}$ on SYSU-MM01 dataset.}
\begin{adjustbox}{max width=0.48\textwidth}
\footnotesize
\begin{tabular}{c|ccc|ccc}
\hline 
\multicolumn{1}{c|}{
\multirow{2}{*}{Value of $k_{te}$}}
& \multicolumn{3}{c|}{All-Search} & \multicolumn{3}{c}{Indoor-Search} \\ 
\cline{2-7}
& R1 & mAP & mINP & R1 & mAP & mINP \\
\hline
$k_{te}=10$ & 76.56 & 73.10 & 60.51 & 78.05 & 81.58 & 78.48 \\
$k_{te}=20$ & 78.00 & 73.94 & 60.67 & 79.14 & 82.44 & 79.29 \\
$k_{te}=30$ & \textbf{78.29} & \textbf{74.08} & \textbf{60.68} & \textbf{79.64} & \textbf{82.75} & \textbf{79.65} \\
$k_{te}=40$ & 77.57 & 73.39 & 59.88 & 79.50 & 82.54 & 79.36 \\
$k_{te}=50$ & 76.56 & 72.34 & 58.68 & 78.59 & 81.79 & 78.47 \\
\hline
\end{tabular}
\end{adjustbox}
\label{tab:k-te-analysis}
\vspace{-4mm}
\end{table}

\begin{table}[!htbp]
\centering	
\renewcommand\arraystretch{1.1}
{\caption{Parameter analysis of $k_{te}$ on RegDB dataset.}}
\begin{adjustbox}{max width=0.48\textwidth}
\footnotesize
\begin{tabular}{c|ccc|ccc}
\hline 
\multicolumn{1}{c|}{
\multirow{2}{*}{Value of $k_{te}$}}
& \multicolumn{3}{c|}{Visible-to-Infrared} & \multicolumn{3}{c}{Infrared-to-Visible} \\ 
\cline{2-7}
& R1 & mAP & mINP & R1 & mAP & mINP \\
\hline
$k_{te}=4$ & 92.77 & 93.01 & \textbf{91.44} & 92.82 & 92.84 & \textbf{90.88}\\
$k_{te}=8$ & \textbf{93.01} & \textbf{93.16} & 91.04 & \textbf{94.08} & \textbf{93.46} & 90.73 \\
$k_{te}=15$ & 92.18 & 92.01 & 89.26 & 92.28 & 92.15 & 89.37 \\
$k_{te}=20$ & 90.53 & 89.09 & 84.15 & 90.78 & 89.51 & 84.61 \\
$k_{te}=25$ & 88.69 & 86.49 & 80.80 & 89.37 & 87.45 & 80.96\\
\hline
\end{tabular}
\end{adjustbox}
\label{tab:k-te-analysis-regdb}
\vspace{-4mm}
\end{table}

\begin{figure}[h]
\begin{minipage}{1.0\linewidth}
\centerline{\includegraphics[width=1.0\textwidth]{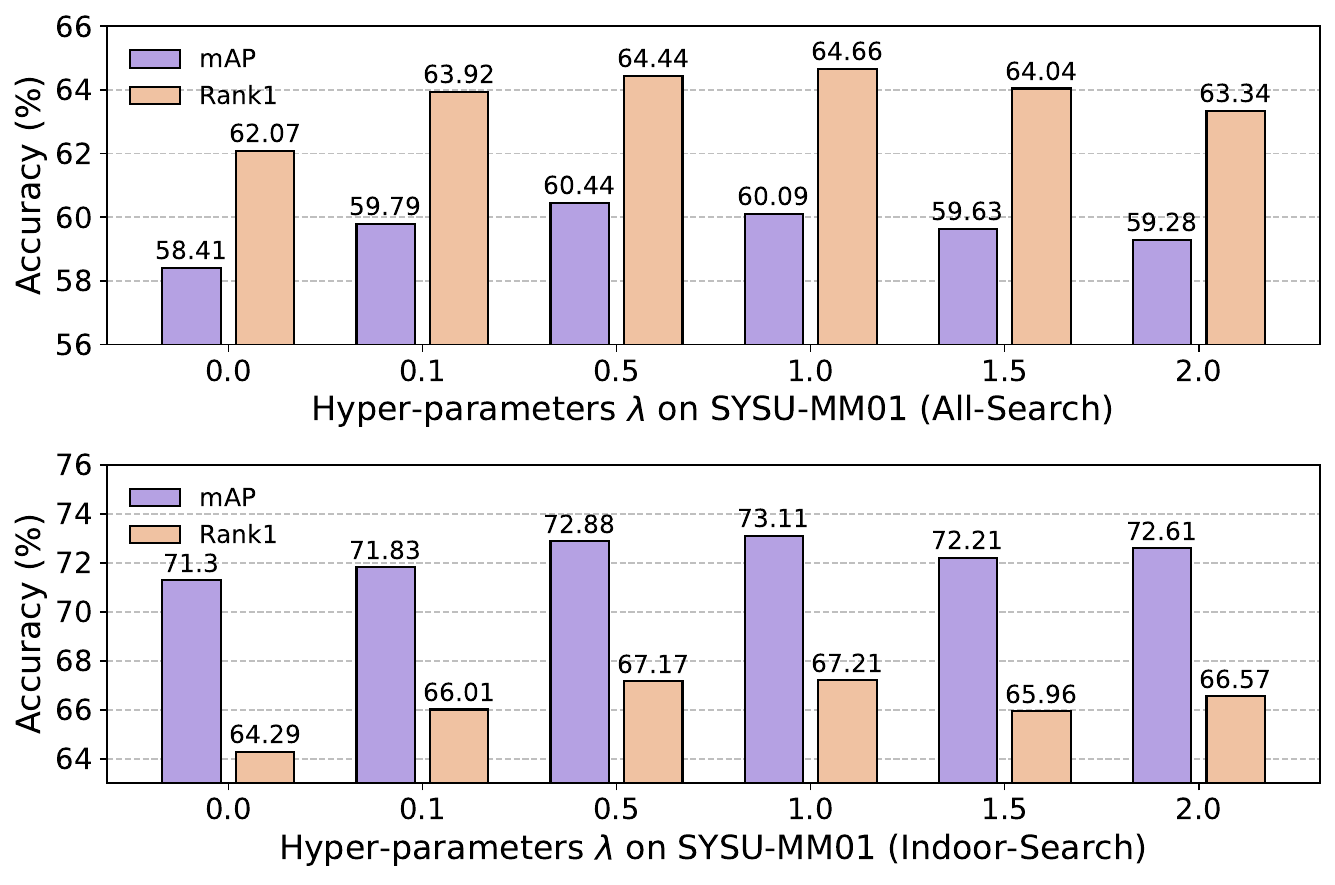}}
\vspace{0mm}
\centerline{\includegraphics[width=1.0\textwidth]{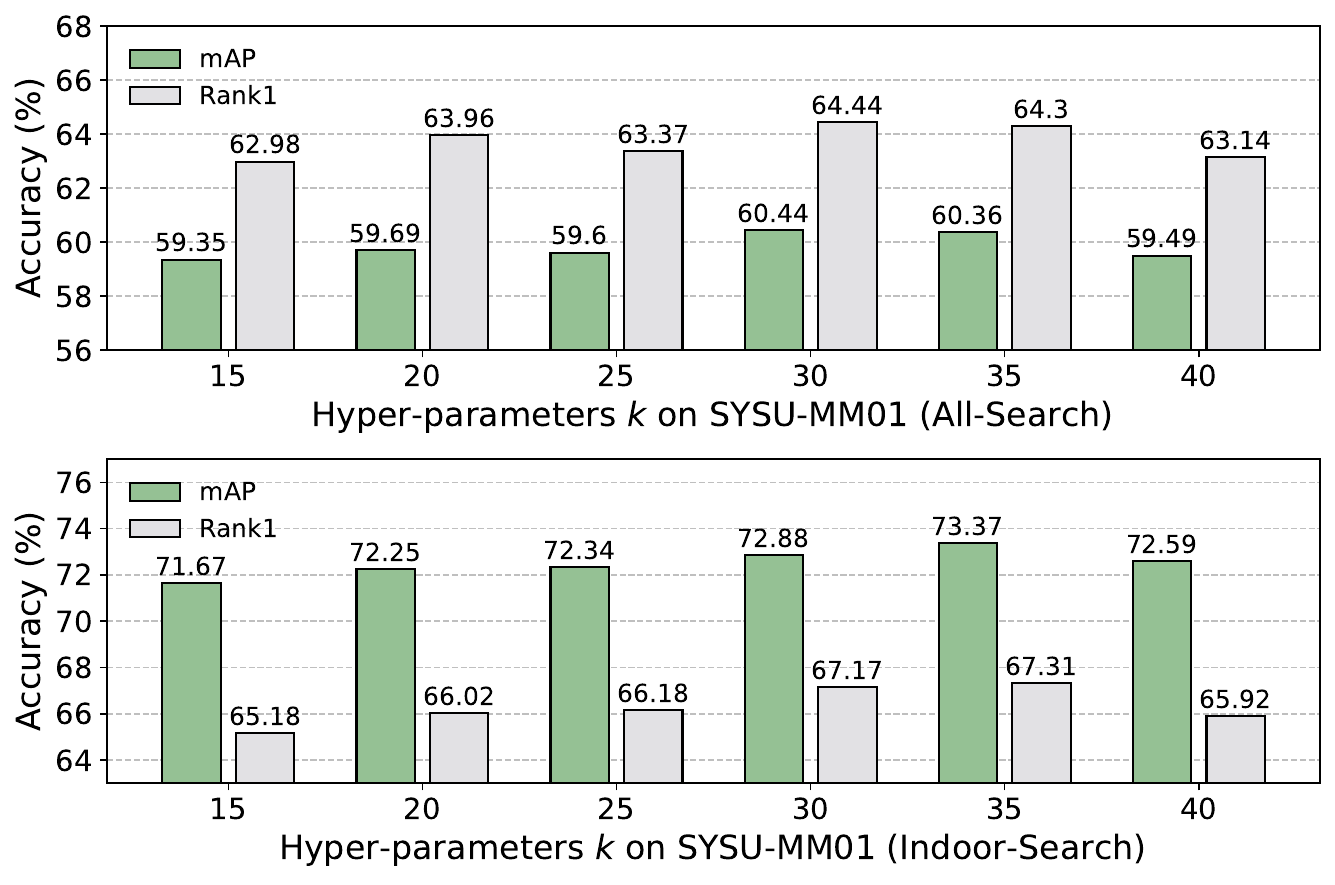}}
\end{minipage}
\vspace{-2mm}
\caption{Parameter analysis of $\lambda$ and $k$ for GPCR on SYSU-MM01 dataset.}
\label{fig:k_lambda_analysis}
\vspace{-6mm}
\end{figure}

\begin{figure}[h]
\begin{minipage}{1.0\linewidth}
\centerline{\includegraphics[width=1.0\textwidth]{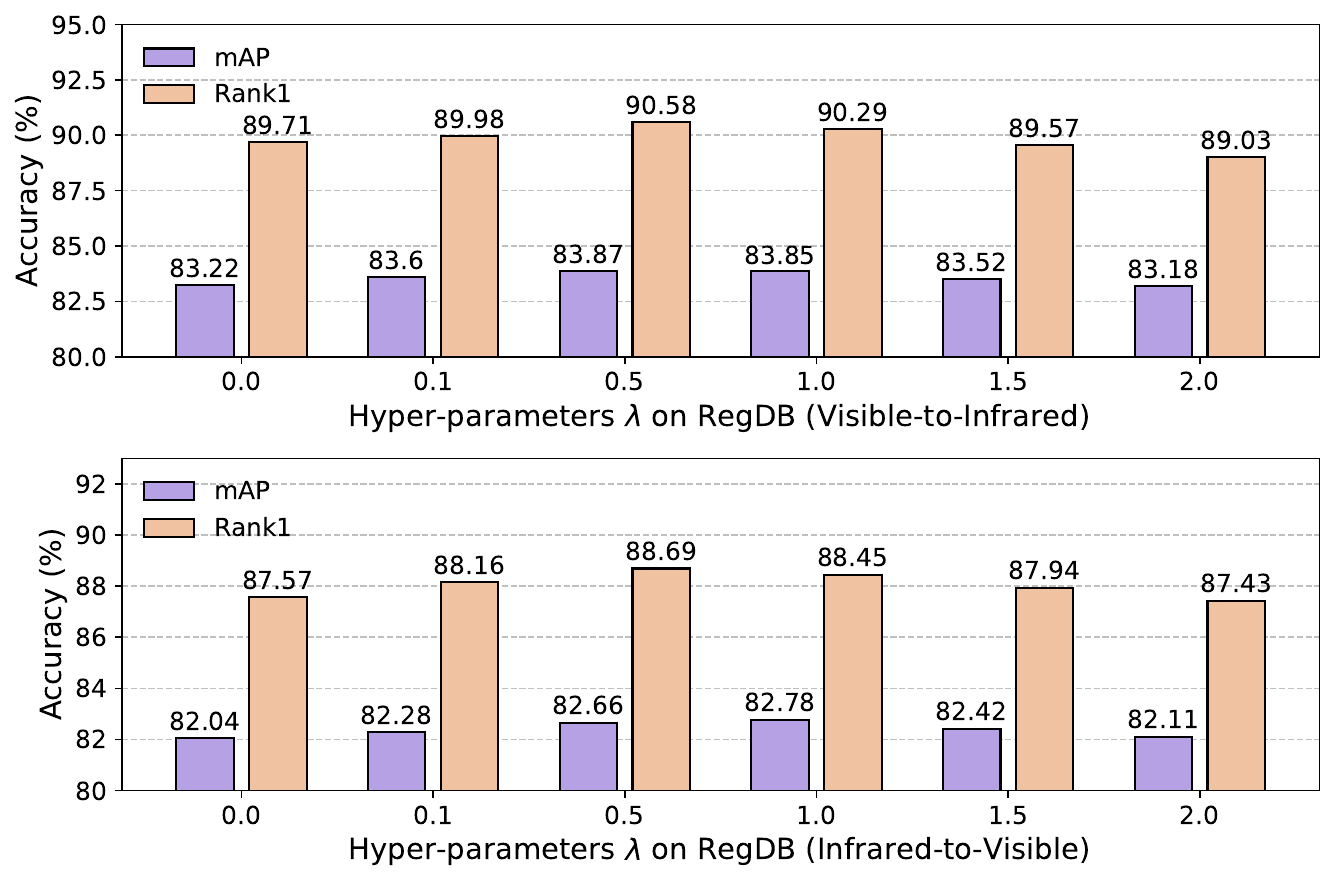}}
\vspace{0mm}
\centerline{\includegraphics[width=1.0\textwidth]{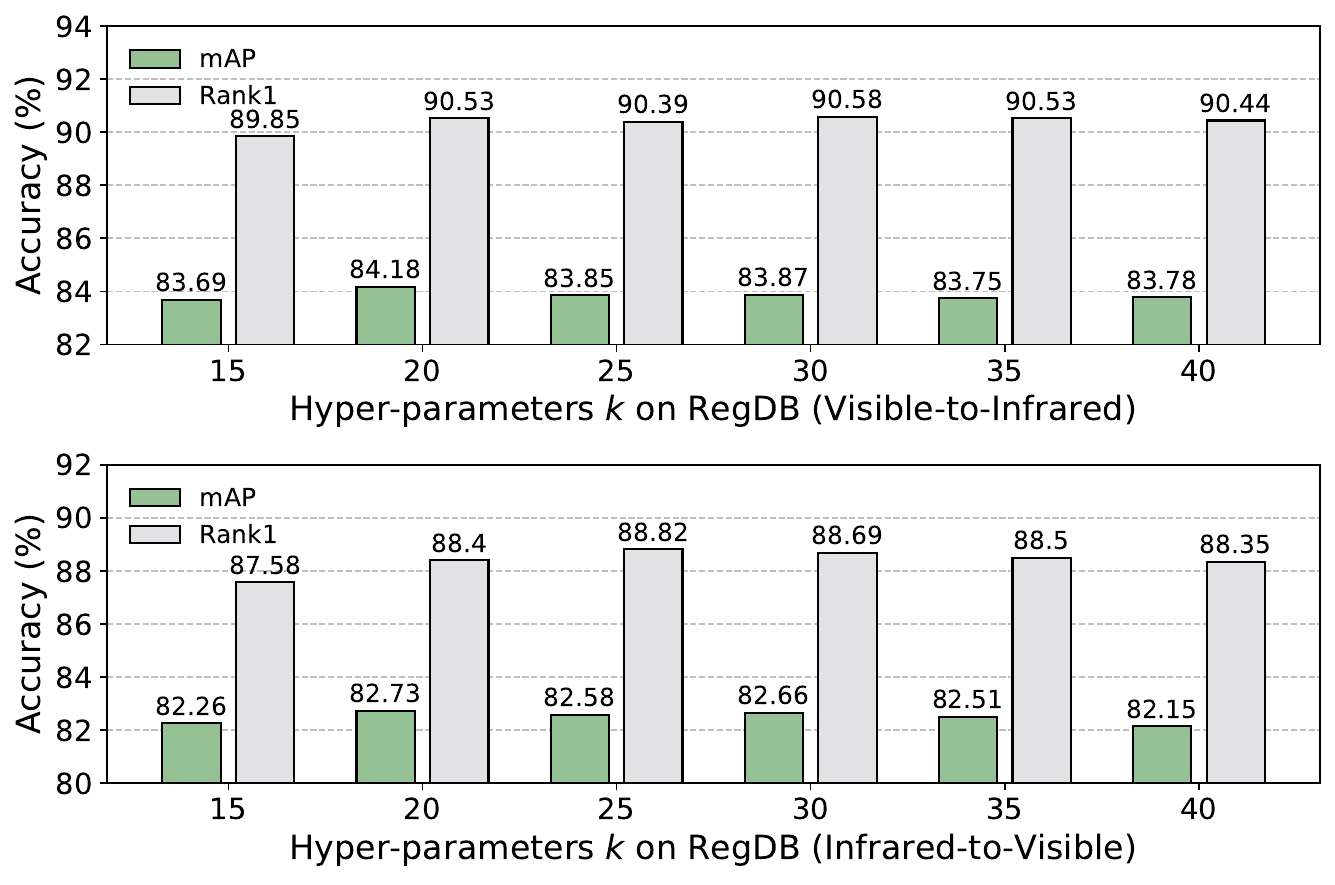}}
\end{minipage}
\vspace{-2mm}
\textcolor{black}{\caption{Parameter analysis of $\lambda$ and $k$ for GPCR on RegDB.}}
\label{fig:k_lambda_analysis_regdb}
\vspace{-4mm}
\end{figure}

\begin{figure}[h]
\centering
\includegraphics[width=0.48\textwidth]{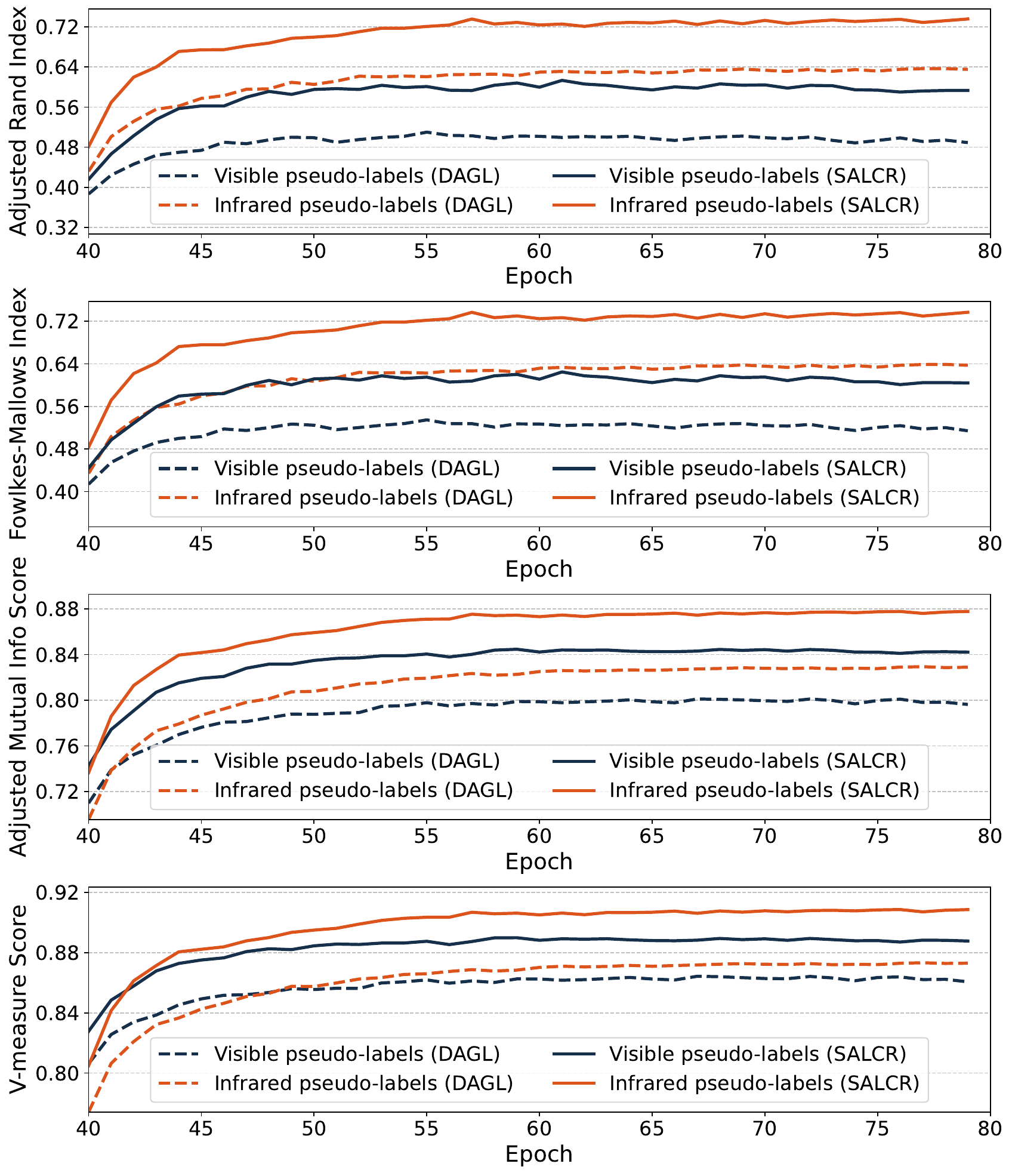}
\caption{Pseudo-label quality analysis over different epochs on SYSU-MM01.}\label{fig:label_quality}
\vspace{-3mm}
\end{figure}

\begin{figure}
\centering
\includegraphics[width=0.48\textwidth]{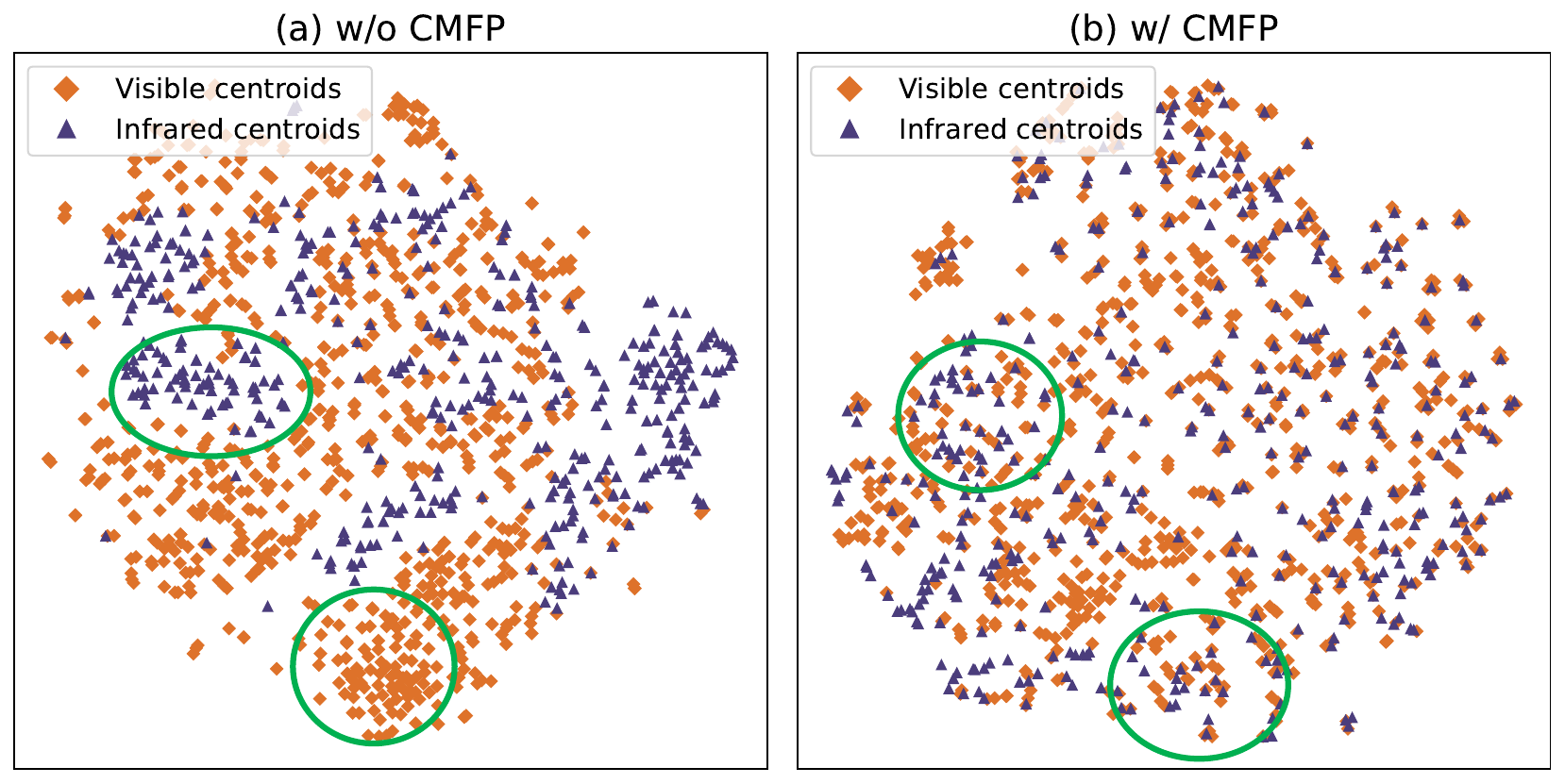}
\caption{Comparison of the cluster centroid distribution before and after CMFP during dual association.}\label{fig:tsne_cmfp}
\vspace{-3mm}
\end{figure}

\subsection{Hyper-Parameter Analysis}

\emph{(a) Hyper-Parameters of FGSAL and GPCR:} 
Our FGSAL and GPCR involve three hyper-parameters:
the number of parts $N_p$, the trade-off parameter $\lambda$ for loss terms and the number of neighbors $k$ for searching positive sets. 
{We analysis these hyper-parameters on SYSU-MM01 and RegDB to study how they behave across different datasets.
Tab.\ref{tab:N-p-analysis} and  shows the effect of $N_p$ in FGSAL. 
An appropriate $N_p$ strikes a trade-off between capturing sufficient fine-grained part information and preserving the semantic integrity of each part. 
When $N_p=3$, the model achieves the best performance on SYSU-MM01 and RegDB.
We consider $N_p=3$ a general solution for most ReID datasets, as dividing the human body into three parts (head, upper body, and lower body) is an intuitive and reasonable strategy.
The effect of $\lambda$ and $k$ in GPCR is illustrated in Fig.\ref{fig:k_lambda_analysis} and Fig.\ref{fig:k_lambda_analysis_regdb}.}
When $\lambda=0$, the GPCR module is disabled, resulting in unsatisfied performance.
The best performance appears when $\lambda=0.5$, where the offline cluster-guided learning and the online instance-guided learning attain a balance and promote each other. 
\textcolor{black}{Notably, too large $\lambda$ ($\lambda=2.0$ in Fig.\ref{fig:k_lambda_analysis_regdb}) may result in a slight performance degradation on RegDB.
It can be attributed to the sufficiently accurate offline pseudo-labels in RegDB, where an excessive reliance on instance-level constraints may result in less compact cluster-level representations.}
The $k$ is utilized to search neighbor sets in the instance memory. Too small $k$ leads to discarding valuable contextual information in the embedding space, while too large $k$ will excessively introduce hard negatives into contrastive learning.
We set $k=30$ based on the experiment results.

\emph{(b) Hyper-Parameters of CMFP:} The CMFP module involve two hyper-parameters $k_{tr}$ and $k_{te}$, corresponding to its employment during training and testing. The effect of $k_{tr}$ and $k_{te}$ are shown in Tab.\ref{tab:k-tr-analysis},  Tab.\ref{tab:k-tr-analysis-regdb}, Tab.\ref{tab:k-te-analysis} and Tab.\ref{tab:k-te-analysis-regdb}. 
The larger value of $k_{tr}$ or $k_{te}$ may introduce an excessive number of negative instances while the smaller value cannot fully exploit the neighborhood relationships. 
\textcolor{black}{When $20\leq k_{tr} \leq 50$, the model achieves satisfactory results on SYSU-MM01, while the optimal range on RegDB is $8 \leq k_{tr} \leq 20$.
The optimal values of $k_{tr}$ and $k_{te}$ are influenced by the dataset scale. For a larger dataset, it is preferable to select relatively larger values for these two hyperparameters.
Based on the results, we set $k_{tr}=30$, $k_{te}=30$ for SYSU-MM01 and $k_{tr}=8$, $k_{te}=8$ for RegDB. 
}
Notably, the gallery set contains only 301 images for retrieval on SYSU-MM01, with an average of 3 images per identity.
When $k_{te}$ is larger than 3, the CMFP inevitably incorporates negative features into the propagation process.
However, as shown in Tab.\ref{tab:k-te-analysis}, the performance is robust when $k_{te}$ varies from 10 to 30, indicating that the structural information brought by negative instances also advances the retrieval performance. 
The results further demonstrate the robustness and applicability of the proposed CMFP.

\begin{figure*}
\centering
\includegraphics[width=1.0\textwidth]{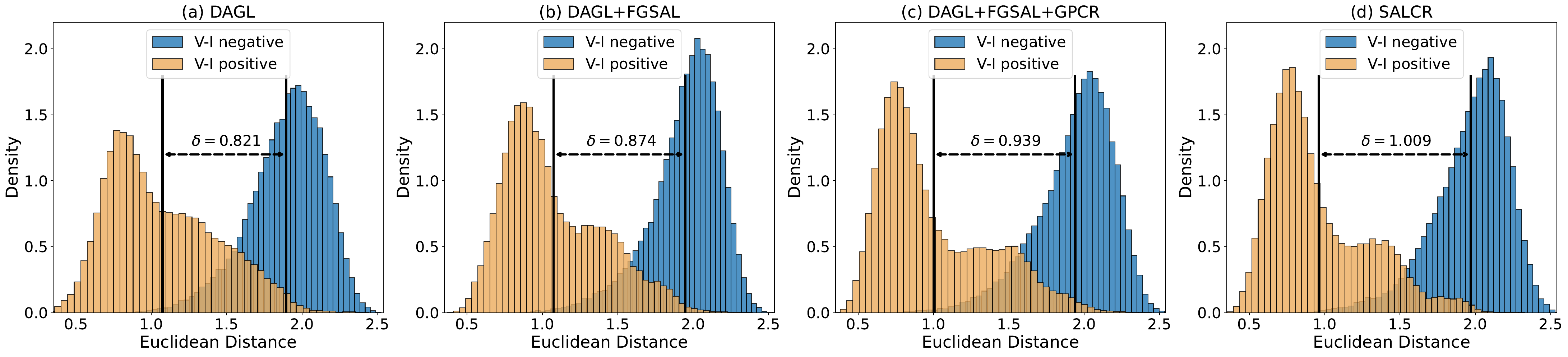}
\caption{The visualization of distance distribution from randomly selected cross-modality positive and negative pairs.}\label{fig:distribution}
\vspace{-3mm}
\end{figure*}

\begin{figure}
\centering
\includegraphics[width=0.48\textwidth]{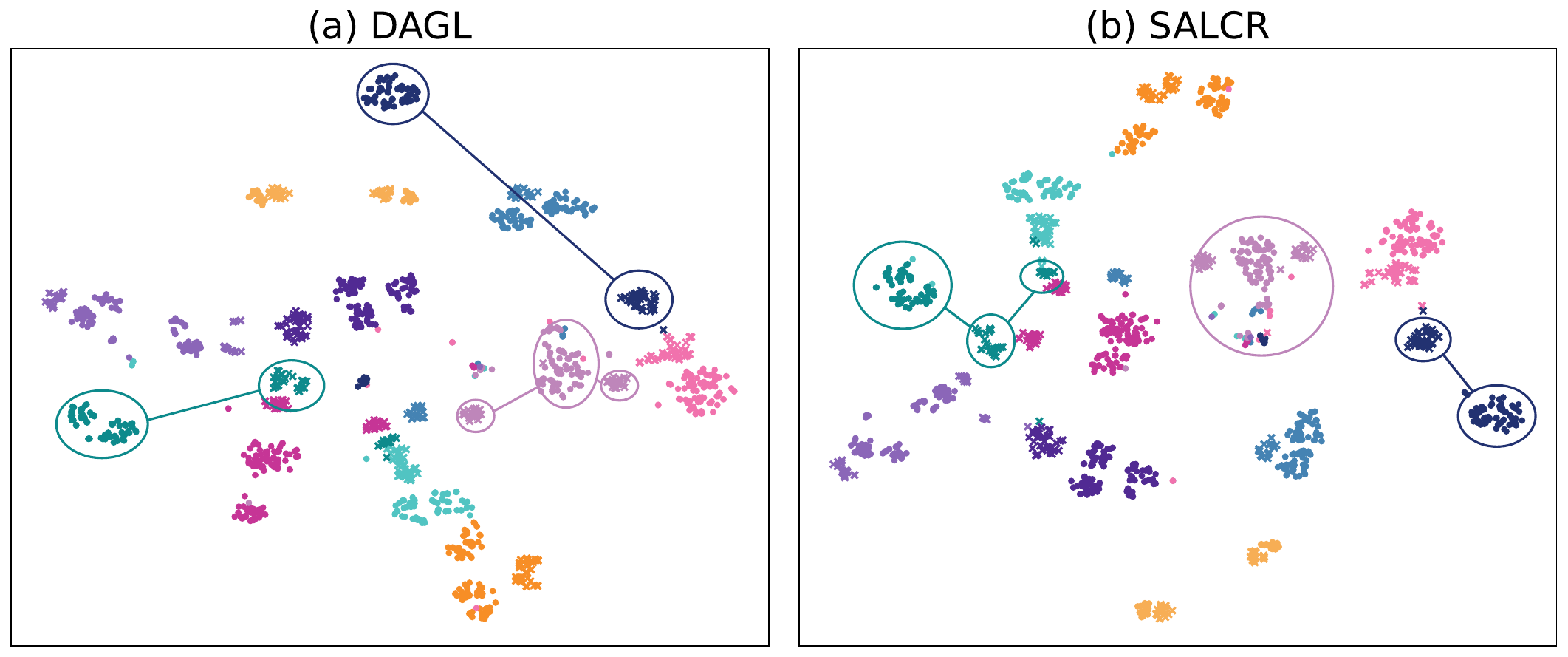}
\caption{The t-SNE visualization of randomly selected identities. Different colors denote different identities. Circles represent instances from the visible modality and crosses represent instances from the infrared modality.}\label{fig:t-sne-ablation}
\vspace{-3mm}
\end{figure}

\begin{figure}
\centering
\includegraphics[width=0.45\textwidth]{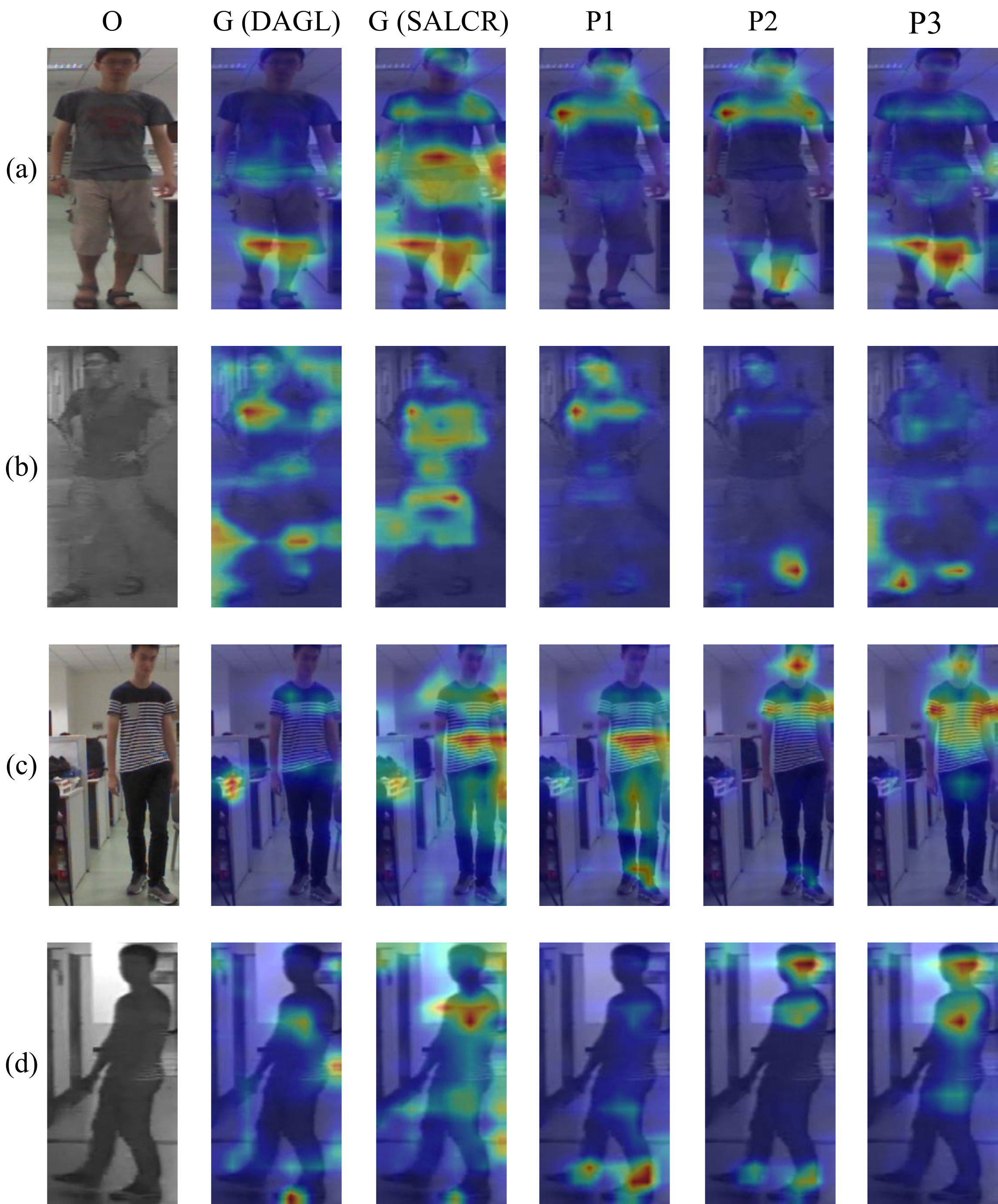}
\caption{The visualization of attention maps from DAGL and SALCR.}\label{fig:grad_cam}
\vspace{-5mm}
\end{figure}

\subsection{Further Analysis}

\emph{(a) Analysis of the quality of the pseudo-labels:}
We employ four clustering quality metrics to evaluate the quality of the pseudo-labels of our method following \cite{ISE}, namely Adjusted Rand Index (ARI), Fowlkes-Mallows Index (FMI), Adjusted Mutual Info Score (AMI) and V-measure score, as shown in Fig.\ref{fig:label_quality}.
The larger score represents the better alignment between the pseudo-labels and the ground-truth labels, signifying the better quality of the pseudo-labels. 
We compare our entire SALCR framework with DAGL in Sec\ref{DAGI}, which solely relies on global features for contrastive learning. 
In Fig.\ref{fig:label_quality}, ``visible pseudo-labels'' represents the metric score between the pseudo-labels for both visible and infrared instances in visible space $\{\tilde{y}^v_i\}_{i=1}^{N^v} \cup \{\hat{y}^r_i\}_{i=1}^{N^r}$ and the ground-truth labels.
The results highlight the improvement of our method in the quality of the pseudo-labels.
The improvement in the early training stages is mainly brought by the CMFP module, which integrates structural information into the embedding space, thereby enhancing the accuracy of dual association.
As the training process goes on, our FGSAL and GPCR modules progressively enhance the pseudo-label quality by learning fine-grained modality-shared information at both the cluster and instance levels.

\emph{(b) Analysis of the effectiveness of CMFP:} We visualize the T-SNE map \cite{t-sne} of the cluster centroids at the 40-th epoch on SYSU-MM01 before and after applying CMFP, as shown in Fig.\ref{fig:tsne_cmfp}. 
Due to the substantial modality discrepancy, instances from different modalities tend to be distributed in distinct manifolds.
Thus the performance of the conventional association methods is limited. 
As illustrated by the green circles in Fig. \ref{fig:tsne_cmfp}(a), the centroids of different clusters within the same modality appear to be gathered together, resulting in unreliable distance metrics for the association.
Such a modality misalignment issue is alleviated when incorporating our CMFP.
As illustrated in Fig. \ref{fig:tsne_cmfp}(b), clusters from different modalities are relatively uniformly distributed in the embedding space. 
The results demonstrate the efficacy of CMFP in bridging the modality gap.

\begin{figure*}[h]
\centering
\includegraphics[width=1.0\textwidth]{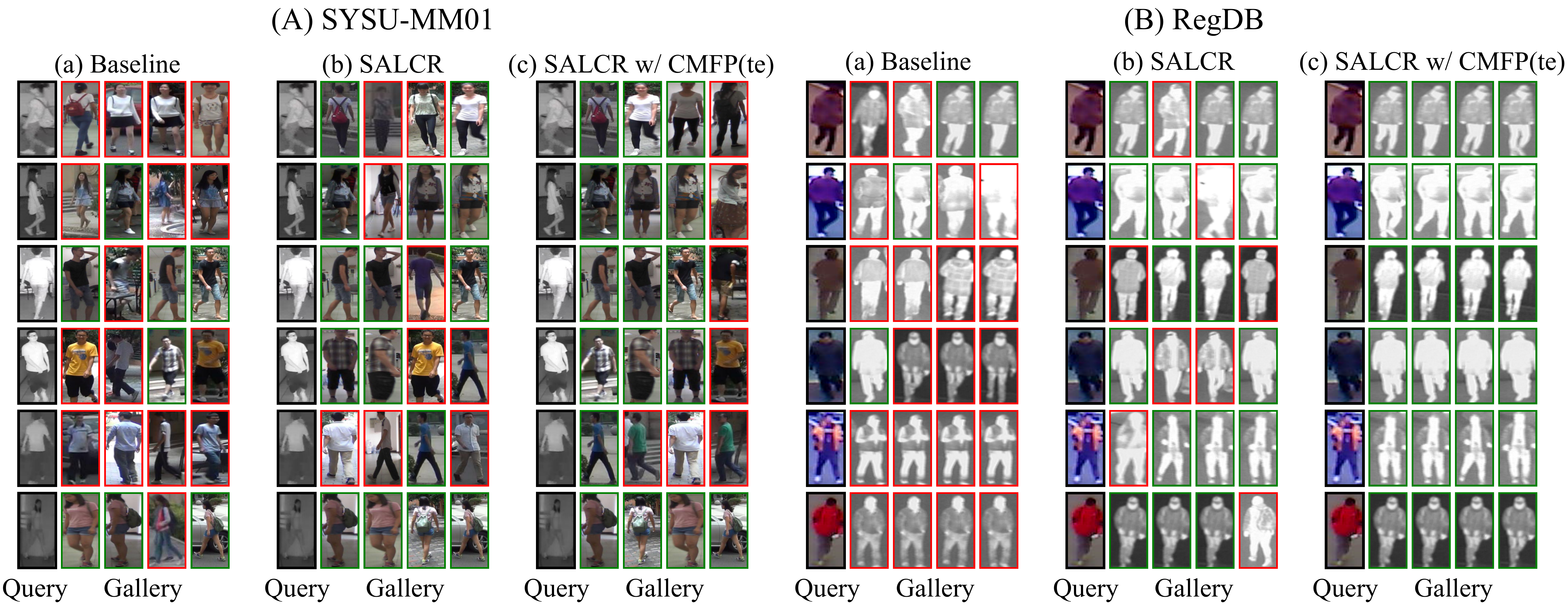}
\caption{Visualization of the ranking lists on the SYSU-MM01 and RegDB datasets. 
The query images are marked with black boxes.
The persons who are different from the query persons are marked with red boxes, while those who are the same as the query are marked with green boxes.}\label{fig:ranking-list}
\vspace{-3mm}
\end{figure*}

\emph{(c) Visualization of the modality discrepancy.} To illustrate the effectiveness of our method in alleviating the modality gap, we visualize the distribution of Euclidean Distance between randomly selected 60,000 positive and negative cross-modality pairs, as shown in Fig.\ref{fig:distribution}. 
With the integration of components in our methods ($i.e.$, FGSAL, GPCR, and CMFP), the peak of the distance distribution for positive pairs gradually shifts leftwards, while the gap between the peak corresponding to positive pairs and that to negative pairs gradually increases.
We further visualize the t-SNE \cite{t-sne} map of randomly selected 11 identities, as shown in Fig.\ref{fig:t-sne-ablation}.
We compare our SALCR with DAGL (Sec.\ref{DAGI}) to demonstrate the effectiveness of part-level information for learning modality-invariant features.
The results indicate that DAGL can learn compact cross-modality features for certain identities with global features, while there are still mismatches in Fig.\ref{fig:t-sne-ablation}.
Our SALCR further narrows the gap between intra-identity features from different modalities, while adjusting some erroneous matches.

\emph{(d) Visualization of the part features:} We visualize the feature map utilizing the Grad-Cam \cite{grad-cam} algorithm, as shown in \ref{fig:grad_cam}.
Specifically, we take the probability prediction from the memory bank constructed by ground-truth labels as input and derive the attention map of the last layer in ResNet-50.
``O'' denotes the original images.
``G (DAGL)'' denotes the attention maps of global features extracted from our OTLA-based baseline model and ``G (SALCR)'' denotes the maps of global features from our SALCR.
``P1'', ``P2'', ``P3'' denotes the attention maps from corresponding part features. 
The results indicate that our method enables the global features to focus on richer fine-grained information, while each part feature concentrates on distinct regions.

\emph{(e) Visualization of Ranking List:}
We visualize the ranking lists of several query images on SYSU-MM01 and RegDB, as shown in Fig.\ref{fig:ranking-list}.
We compare our SALCR and SALCR w/ CMFP(te) with the baseline trained under the DCL framework \cite{ADCA}.
The results demonstrate the effectiveness of our SALCR framework for learning fine-grained modality-shared information during training, and our CMFP for leveraging neighborhood information during testing.

\textcolor{black}{\emph{(f) Analysis of Parameter Count and Computational Overhead:}
We analyze the additional number of parameters and the computational overhead introduced by the part feature generation, and the results are shown in Tab.\ref{tab:parameter-analysis}.
Our part feature generation process adds only 0.35M (1.36\%) in additional parameters and 0.01 GMac in extra FLOPs. 
However, it results in a performance improvement of 5.04\% mAP compared to the approach using only global features (DAGL+CMFP(tr), Idx 6 in Tab.\ref{tab:ablation}).
The results demonstrate that our part feature significantly improves model performance with almost no additional computational overhead.
}

\begin{table}[!htbp]
\vspace{-3mm}
\centering	
\renewcommand\arraystretch{1.1}
\textcolor{black}{\caption{Analysis of the parameter count and computational overhead.}}
\begin{adjustbox}{max width=0.48\textwidth}
\footnotesize
\begin{tabular}{c|c|c|ccc}
\hline 
\multirow{2}*{Models} & \multirow{2}*{Params (M)} & \multirow{2}*{Flops (GMac)} & \multicolumn{3}{c}{SYSU-MM01} \\
\cline{4-6}
& & & R1 & mAP & mINP \\
\hline
ResNet-50 & 25.57(\textcolor{black}{+0.00}) & 5.19(\textcolor{black}{+0.00}) & 60.12 & 55.38 & 42.64 \\
FGSAL ($N_p=3$) & 25.92 (\textcolor{black}{+0.35}) & 5.20(\textcolor{black}{+0.01}) & \textbf{64.44} & \textbf{60.44} & \textbf{45.19}\\
\hline
\end{tabular}
\end{adjustbox}
\label{tab:parameter-analysis}
\vspace{-6mm}
\end{table}



\begin{table}[!htbp]
\vspace{-4mm}
\centering	
\renewcommand\arraystretch{1.1}
\textcolor{black}{\caption{The performance of models trained on different datasets}}
\begin{adjustbox}{max width=0.48\textwidth}
\footnotesize
\begin{tabular}{c|ccc|ccc}
\hline
\multicolumn{1}{c|}{
\multirow{2}{*}{Dataset}}
& \multicolumn{3}{c|}{SYSU-MM01} & \multicolumn{3}{c}{RegDB (seed=1)} \\ 
\cline{2-7}
& R1 & mAP & mINP & R1 & mAP & mINP \\
\hline
SYSU-MM01 & \textbf{64.44}	& \textbf{60.44} &	\textbf{45.19}	& 7.86	& 8.40	& 4.96\\
RegDB (seed=1) & 3.83	& 5.57	& 2.29	& 90.97	& \textbf{85.02}	& \textbf{72.90}\\
Unified dataset & 52.46	& 50.09	& 35.92	& \textbf{94.22}	& 76.29	& 50.90\\
\hline
\end{tabular}
\end{adjustbox}
\label{tab:analysis-generaizability}
\vspace{-6mm}
\end{table}

\begin{table}[!htbp]
\vspace{-5mm}
\centering	
\renewcommand\arraystretch{1.1}
\textcolor{black}{\caption{Parameter analysis of $k_{te}$ with the unified model.}}
\begin{adjustbox}{max width=0.48\textwidth}
\footnotesize
\begin{tabular}{c|ccc|ccc}
\hline 
\multicolumn{1}{c|}{
\multirow{2}{*}{Value of $k_{te}$}}
& \multicolumn{3}{c|}{SYSU-MM01} & \multicolumn{3}{c}{RegDB} \\ 
\cline{2-7}
& R1 & mAP & mINP & R1 & mAP & mINP \\
\hline
$k_{te}=0$ (w/o CMFP(te)) & 52.46	& 50.09	& 35.92	& 94.22	& 76.29	& 50.90 \\
$k_{te}=5$ & 59.04	& 58.15	& 45.84	& 95.31	& \textbf{89.35}	& \textbf{78.95} \\
$k_{te}=10$ & 60.23	& 59.05	& 46.32	& \textbf{95.53}	& 88.40	& 76.57 \\
$k_{te}=20$ & \textbf{61.26}	& \textbf{59.14}	& \textbf{46.63}	&93.64	& 79.87	& 60.66\\
$k_{te}=30$ & 60.30	& 58.33	& 45.07	& 91.70	& 74.04	& 50.33\\
\hline
\end{tabular}
\end{adjustbox}
\label{tab:unified-kte}
\vspace{-5mm}
\end{table}

\textcolor{black}{
\emph{(h) Analysis of the model generalizability:}
To further investigate the generalizability of our framework, we combine the training sets of two USL-VI-ReID datasets ($i.e.$, SYSU-MM01 and RegDB) to train a unified model and evaluate it separately on the corresponding test sets of the two datasets. 
Specifically, the unified training set consists of the training set of SYSU-MM01 and the training set split from RegDB with random seed 1.
We compare the performance of the unified model with models individually trained on each dataset. 
The results, presented in Tab.\ref{tab:analysis-generaizability}, reveal that models trained on a single dataset perform poorly when tested on the other dataset. 
This performance drop is largely due to the significant domain gap between the two datasets. 
Surprisingly, the global model trained on the unified dataset achieves competitive performance on both test domains, despite the large domain differences. 
These findings further highlight the scalability and generalizability of our framework.
We also analyze the effectiveness of CMFP on the unified model, as shown in Tab.\ref{tab:unified-kte}.
The results reveal that CMFP continues to significantly improve the performance of the unified model across different test domains.
}

\subsection{Conclusion}

In this paper, we propose a Semantic-Aligned Learning with Collaborative Refinement (SALCR) framework for USL-VI-ReID. Specifically, 
we first introduce a Dual Association with Global Learning (DAGL) module, which establishes bi-directional cross-modality label associations and constructs a contrastive learning framework for global features.
Then we devise a Fine-Grained Semantic-Aligned Learning (FGSAL) module to learn fine-grained modality-shared information from semantic-aligned pairs, thereby achieving complementarity between label distributions across different modalities. 
To mitigate the side-effects of noisy pseudo-labels, we propose a Global-Part Collaborative Refinement (GPCR) module, which dynamically mines positive sample sets for global and part features.
Moreover, we propose a Cross-Modality Feature Propagation (CMFP) module to enhance the cross-modality relationships during training and testing, achieving a trade-off between efficiency and effectiveness.
Extensive experiments demonstrate the effectiveness of our proposed method, pushing USL-VI-ReID to real applications.

\noindent\textbf{Data Availability Statement} 

\noindent\textbf{SYSU-MM01:} A signed dataset release
\href{https://github.com/wuancong/SYSU-MM01}{\textcolor{blue}{agreement}} must be send to \href{mailto:wuancong@gmail.com}{\textcolor{blue}{wuancong@gmail.com}} or \href{mailto:wuanc@mail.sysu.edu.cn}{\textcolor{blue}{wuanc@mail.sysu.edu.cn}} to obtain a download link.

\noindent\textbf{RegDB:} The dataset can be downloaded by submitting a copyright form to \href{http://dm.dongguk.edu/link.html}{\textcolor{blue}{http://dm.dongguk.edu/link.html}}.

\bibliographystyle{spbasic}
\bibliography{sn-bibliography}    

\begin{thebibliography}{69}
\providecommand{\natexlab}[1]{#1}
\providecommand{\url}[1]{{#1}}
\providecommand{\urlprefix}{URL }
\expandafter\ifx\csname urlstyle\endcsname\relax
  \providecommand{\doi}[1]{DOI~\discretionary{}{}{}#1}\else
  \providecommand{\doi}{DOI~\discretionary{}{}{}\begingroup \urlstyle{rm}\Url}\fi
\providecommand{\eprint}[2][]{\url{#2}}

\bibitem[{Alehdaghi et~al(2022)Alehdaghi, Josi, Cruz, and Granger}]{LUPI}
Alehdaghi M, Josi A, Cruz RM, Granger E (2022) Visible-infrared person re-identification using privileged intermediate information. In: European Conference on Computer Vision, Springer, pp 720--737

\bibitem[{Chen et~al(2021)Chen, Lagadec, and Bremond}]{ICE}
Chen H, Lagadec B, Bremond F (2021) Ice: Inter-instance contrastive encoding for unsupervised person re-identification. In: Proceedings of the IEEE/CVF International Conference on Computer Vision, pp 14960--14969

\bibitem[{Chen et~al(2023)Chen, Zhang, Tan, Qu, and Xie}]{CCLNet}
Chen Z, Zhang Z, Tan X, Qu Y, Xie Y (2023) Unveiling the power of clip in unsupervised visible-infrared person re-identification. In: Proceedings of the 31st ACM International Conference on Multimedia, Association for Computing Machinery, New York, NY, USA, MM '23, p 3667–3675, \doi{10.1145/3581783.3612050}, \urlprefix\url{https://doi.org/10.1145/3581783.3612050}

\bibitem[{Cheng et~al(2023{\natexlab{a}})Cheng, He, Wang, Zhang, Wang, and Gao}]{MBCCM}
Cheng D, He L, Wang N, Zhang S, Wang Z, Gao X (2023{\natexlab{a}}) Efficient bilateral cross-modality cluster matching for unsupervised visible-infrared person reid. In: Proceedings of the 31st ACM International Conference on Multimedia, pp 1325--1333

\bibitem[{Cheng et~al(2023{\natexlab{b}})Cheng, Huang, Wang, He, Li, and Gao}]{DOTLA}
Cheng D, Huang X, Wang N, He L, Li Z, Gao X (2023{\natexlab{b}}) Unsupervised visible-infrared person reid by collaborative learning with neighbor-guided label refinement. In: Proceedings of the 31st ACM International Conference on Multimedia, pp 7085--7093

\bibitem[{Cho et~al(2022)Cho, Kim, Hong, and Yoon}]{PPLR}
Cho Y, Kim WJ, Hong S, Yoon SE (2022) Part-based pseudo label refinement for unsupervised person re-identification. In: Proceedings of the IEEE/CVF conference on computer vision and pattern recognition, pp 7308--7318

\bibitem[{Choi et~al(2020)Choi, Lee, Kim, Kim, and Kim}]{hi-cmd}
Choi S, Lee S, Kim Y, Kim T, Kim C (2020) Hi-cmd: Hierarchical cross-modality disentanglement for visible-infrared person re-identification. In: Proceedings of the IEEE/CVF conference on computer vision and pattern recognition, pp 10257--10266

\bibitem[{Cuturi(2013)}]{sinkhorn}
Cuturi M (2013) Sinkhorn distances: Lightspeed computation of optimal transport. Advances in neural information processing systems 26

\bibitem[{Dai et~al(2018)Dai, Ji, Wang, Wu, and Huang}]{cmGAN}
Dai P, Ji R, Wang H, Wu Q, Huang Y (2018) Cross-modality person re-identification with generative adversarial training. In: IJCAI, vol~1, p~6

\bibitem[{Dai et~al(2022)Dai, Wang, Yuan, Zhu, and Tan}]{ClusterContrast}
Dai Z, Wang G, Yuan W, Zhu S, Tan P (2022) Cluster contrast for unsupervised person re-identification. In: Proceedings of the Asian Conference on Computer Vision, pp 1142--1160

\bibitem[{Deng et~al(2009)Deng, Dong, Socher, Li, Li, and Fei-Fei}]{ImageNet}
Deng J, Dong W, Socher R, Li LJ, Li K, Fei-Fei L (2009) Imagenet: A large-scale hierarchical image database. In: 2009 IEEE conference on computer vision and pattern recognition, Ieee, pp 248--255

\bibitem[{Ester et~al(1996)Ester, Kriegel, Sander, Xu et~al}]{1996DBSCAN}
Ester M, Kriegel HP, Sander J, Xu X, et~al (1996) A density-based algorithm for discovering clusters in large spatial databases with noise. In: kdd, vol~96, pp 226--231

\bibitem[{Fang et~al(2023)Fang, Yang, and Fu}]{SAAI}
Fang X, Yang Y, Fu Y (2023) Visible-infrared person re-identification via semantic alignment and affinity inference. In: Proceedings of the IEEE/CVF International Conference on Computer Vision, pp 11270--11279

\bibitem[{Feng et~al(2023)Feng, Wu, and Zheng}]{SEFEL}
Feng J, Wu A, Zheng WS (2023) Shape-erased feature learning for visible-infrared person re-identification. In: Proceedings of the IEEE/CVF Conference on Computer Vision and Pattern Recognition (CVPR), pp 22752--22761

\bibitem[{Frigo and Johnson(1998)}]{FFT}
Frigo M, Johnson SG (1998) Fftw: An adaptive software architecture for the fft. In: Proceedings of the 1998 IEEE International Conference on Acoustics, Speech and Signal Processing, ICASSP'98 (Cat. No. 98CH36181), IEEE, vol~3, pp 1381--1384

\bibitem[{Ge et~al(2020{\natexlab{a}})Ge, Chen, and Li}]{MMT}
Ge Y, Chen D, Li H (2020{\natexlab{a}}) Mutual mean-teaching: Pseudo label refinery for unsupervised domain adaptation on person re-identification. In: International Conference on Learning Representations, \urlprefix\url{https://openreview.net/forum?id=rJlnOhVYPS}

\bibitem[{Ge et~al(2020{\natexlab{b}})Ge, Zhu, Chen, Zhao et~al}]{SPCL}
Ge Y, Zhu F, Chen D, Zhao R, et~al (2020{\natexlab{b}}) Self-paced contrastive learning with hybrid memory for domain adaptive object re-id. Advances in neural information processing systems 33:11309--11321

\bibitem[{He et~al(2016)He, Zhang, Ren, and Sun}]{resnet}
He K, Zhang X, Ren S, Sun J (2016) Deep residual learning for image recognition. In: Proceedings of the IEEE conference on computer vision and pattern recognition, pp 770--778

\bibitem[{He et~al(2020)He, Fan, Wu, Xie, and Girshick}]{moco}
He K, Fan H, Wu Y, Xie S, Girshick R (2020) Momentum contrast for unsupervised visual representation learning. In: Proceedings of the IEEE/CVF conference on computer vision and pattern recognition, pp 9729--9738

\bibitem[{He et~al(2024)He, Cheng, Wang, and Gao}]{he2024exploring}
He L, Cheng D, Wang N, Gao X (2024) Exploring homogeneous and heterogeneous consistent label associations for unsupervised visible-infrared person reid. International Journal of Computer Vision pp 1--20

\bibitem[{Jiang et~al(2022)Jiang, Zhang, Liu, Qian, Zhang, and Wu}]{CMT}
Jiang K, Zhang T, Liu X, Qian B, Zhang Y, Wu F (2022) Cross-modality transformer for visible-infrared person re-identification. In: European Conference on Computer Vision, Springer, pp 480--496

\bibitem[{Kim et~al(2023)Kim, Kim, Park, Park, and Sohn}]{PartMix}
Kim M, Kim S, Park J, Park S, Sohn K (2023) Partmix: Regularization strategy to learn part discovery for visible-infrared person re-identification. In: Proceedings of the IEEE/CVF Conference on Computer Vision and Pattern Recognition (CVPR), pp 18621--18632

\bibitem[{Li et~al(2022)Li, Lu, Liu, Liu, Yin, Chu, Huang, Zhu, Zhao, and Yu}]{CIFT}
Li X, Lu Y, Liu B, Liu Y, Yin G, Chu Q, Huang J, Zhu F, Zhao R, Yu N (2022) Counterfactual intervention feature transfer for visible-infrared person re-identification. In: European Conference on Computer Vision, Springer, pp 381--398

\bibitem[{Li et~al(2024)Li, Zhang, and Zhang}]{Freq2}
Li Y, Zhang T, Zhang Y (2024) Frequency domain modality-invariant feature learning for visible-infrared person re-identification. arXiv preprint arXiv:240101839

\bibitem[{Liang et~al(2024)Liang, Jin, Liu, Wang, Feng, and Li}]{MCJA}
Liang T, Jin Y, Liu W, Wang T, Feng S, Li Y (2024) Bridging the gap: Multi-level cross-modality joint alignment for visible-infrared person re-identification. IEEE Transactions on Circuits and Systems for Video Technology

\bibitem[{Liang et~al(2021)Liang, Wang, Lai, and Xie}]{H2H}
Liang W, Wang G, Lai J, Xie X (2021) Homogeneous-to-heterogeneous: Unsupervised learning for rgb-infrared person re-identification. IEEE Transactions on Image Processing 30:6392--6407

\bibitem[{Liu et~al(2022)Liu, Sun, Zhu, Pei, Yang, and Li}]{MAUM}
Liu J, Sun Y, Zhu F, Pei H, Yang Y, Li W (2022) Learning memory-augmented unidirectional metrics for cross-modality person re-identification. In: Proceedings of the IEEE/CVF Conference on Computer Vision and Pattern Recognition, pp 19366--19375

\bibitem[{Lu et~al(2020)Lu, Wu, Liu, Zhang, Li, Chu, and Yu}]{cm-SSFT}
Lu Y, Wu Y, Liu B, Zhang T, Li B, Chu Q, Yu N (2020) Cross-modality person re-identification with shared-specific feature transfer. In: Proceedings of the IEEE/CVF Conference on Computer Vision and Pattern Recognition (CVPR)

\bibitem[{Luo et~al(2019)Luo, Chen, Wang, and Zhang}]{SSFT}
Luo C, Chen Y, Wang N, Zhang Z (2019) Spectral feature transformation for person re-identification. In: Proceedings of the IEEE/CVF international conference on computer vision, pp 4976--4985

\bibitem[{Van~der Maaten and Hinton(2008)}]{t-sne}
Van~der Maaten L, Hinton G (2008) Visualizing data using t-sne. Journal of machine learning research 9(11)

\bibitem[{Mao et~al(2017)Mao, Li, and Xie}]{AlignGAN}
Mao X, Li Q, Xie H (2017) Aligngan: Learning to align cross-domain images with conditional generative adversarial networks. \eprint{1707.01400}

\bibitem[{Nguyen et~al(2017)Nguyen, Hong, Kim, and Park}]{RegDB}
Nguyen DT, Hong HG, Kim KW, Park KR (2017) Person recognition system based on a combination of body images from visible light and thermal cameras. Sensors 17(3):605

\bibitem[{Pang et~al(2023)Pang, Wang, Zhao, Liu, and Sharma}]{CHCR}
Pang Z, Wang C, Zhao L, Liu Y, Sharma G (2023) Cross-modality hierarchical clustering and refinement for unsupervised visible-infrared person re-identification. IEEE Transactions on Circuits and Systems for Video Technology pp 1--1, \doi{10.1109/TCSVT.2023.3310015}

\bibitem[{Pang et~al(2024{\natexlab{a}})Pang, Wang, Pan, Zhao, Wang, and Guo}]{MIMR}
Pang Z, Wang C, Pan H, Zhao L, Wang J, Guo M (2024{\natexlab{a}}) Mimr: Modality-invariance modeling and refinement for unsupervised visible-infrared person re-identification. Knowledge-Based Systems 285:111350, \doi{https://doi.org/10.1016/j.knosys.2023.111350}, \urlprefix\url{https://www.sciencedirect.com/science/article/pii/S0950705123010985}

\bibitem[{Pang et~al(2024{\natexlab{b}})Pang, Zhao, Liu, Sharma, and Wang}]{ISML-TCSVT24}
Pang Z, Zhao L, Liu Y, Sharma G, Wang C (2024{\natexlab{b}}) Inter-modality similarity learning for unsupervised multi-modality person re-identification. IEEE Transactions on Circuits and Systems for Video Technology

\bibitem[{Radford et~al(2021)Radford, Kim, Hallacy, Ramesh, Goh, Agarwal, Sastry, Askell, Mishkin, Clark et~al}]{CLIP}
Radford A, Kim JW, Hallacy C, Ramesh A, Goh G, Agarwal S, Sastry G, Askell A, Mishkin P, Clark J, et~al (2021) Learning transferable visual models from natural language supervision. In: International conference on machine learning, PMLR, pp 8748--8763

\bibitem[{Ren and Zhang(2024)}]{IDKL}
Ren K, Zhang L (2024) Implicit discriminative knowledge learning for visible-infrared person re-identification. arXiv preprint arXiv:240311708

\bibitem[{Selvaraju et~al(2017)Selvaraju, Cogswell, Das, Vedantam, Parikh, and Batra}]{grad-cam}
Selvaraju RR, Cogswell M, Das A, Vedantam R, Parikh D, Batra D (2017) Grad-cam: Visual explanations from deep networks via gradient-based localization. In: Proceedings of the IEEE international conference on computer vision, pp 618--626

\bibitem[{Shi et~al(2023)Shi, Zhang, Yin, Xie, Zhang, Fan, Shi, and Qu}]{DPIS}
Shi J, Zhang Y, Yin X, Xie Y, Zhang Z, Fan J, Shi Z, Qu Y (2023) Dual pseudo-labels interactive self-training for semi-supervised visible-infrared person re-identification. In: Proceedings of the IEEE/CVF International Conference on Computer Vision, pp 11218--11228

\bibitem[{Shi et~al(2024{\natexlab{a}})Shi, Yin, Chen, Zhang, Zhang, Xie, and Qu}]{MMM}
Shi J, Yin X, Chen Y, Zhang Y, Zhang Z, Xie Y, Qu Y (2024{\natexlab{a}}) Multi-memory matching for unsupervised visible-infrared person re-identification. \eprint{2401.06825}

\bibitem[{Shi et~al(2024{\natexlab{b}})Shi, Yin, Wang, Liu, Xie, and Qu}]{PCMIP}
Shi J, Yin X, Wang Y, Liu X, Xie Y, Qu Y (2024{\natexlab{b}}) Progressive contrastive learning with multi-prototype for unsupervised visible-infrared person re-identification. arXiv preprint arXiv:240219026

\bibitem[{Si et~al(2023)Si, He, Li, Song, and Fan}]{DFC}
Si T, He F, Li P, Song Y, Fan L (2023) Diversity feature constraint based on heterogeneous data for unsupervised person re-identification. Information Processing \& Management 60(3):103304

\bibitem[{Sun et~al(2018)Sun, Zheng, Yang, Tian, and Wang}]{PCB}
Sun Y, Zheng L, Yang Y, Tian Q, Wang S (2018) Beyond part models: Person retrieval with refined part pooling (and a strong convolutional baseline). In: Proceedings of the European conference on computer vision (ECCV), pp 480--496

\bibitem[{Tan et~al(2023)Tan, Zhang, Shen, Wang, Dai, Lin, Wu, and Ji}]{LTG}
Tan L, Zhang Y, Shen S, Wang Y, Dai P, Lin X, Wu Y, Ji R (2023) Exploring invariant representation for visible-infrared person re-identification. \eprint{2302.00884}

\bibitem[{Teng et~al(2024)Teng, Shen, Xu, and Lan}]{BCGM-MM24}
Teng X, Shen X, Xu K, Lan L (2024) Enhancing unsupervised visible-infrared person re-identification with bidirectional-consistency gradual matching. In: Proceedings of the 32nd ACM International Conference on Multimedia, pp 9856--9865

\bibitem[{Tian et~al(2021)Tian, Zhang, Lin, Qu, Xie, and Ma}]{VCD+VML}
Tian X, Zhang Z, Lin S, Qu Y, Xie Y, Ma L (2021) Farewell to mutual information: Variational distillation for cross-modal person re-identification. In: Proceedings of the IEEE/CVF Conference on Computer Vision and Pattern Recognition, pp 1522--1531

\bibitem[{Vaswani et~al(2017)Vaswani, Shazeer, Parmar, Uszkoreit, Jones, Gomez, Kaiser, and Polosukhin}]{Attention}
Vaswani A, Shazeer N, Parmar N, Uszkoreit J, Jones L, Gomez AN, Kaiser {\L}, Polosukhin I (2017) Attention is all you need. Advances in neural information processing systems 30

\bibitem[{Wang et~al(2022{\natexlab{a}})Wang, Zhang, Chen, Zhang, Wang, Sheng, Qu, and Xie}]{OTLA}
Wang J, Zhang Z, Chen M, Zhang Y, Wang C, Sheng B, Qu Y, Xie Y (2022{\natexlab{a}}) Optimal transport for label-efficient visible-infrared person re-identification. In: European Conference on Computer Vision, Springer, pp 93--109

\bibitem[{Wang et~al(2022{\natexlab{b}})Wang, Li, Lai, Gong, and Hua}]{O2CAP}
Wang M, Li J, Lai B, Gong X, Hua XS (2022{\natexlab{b}}) Offline-online associated camera-aware proxies for unsupervised person re-identification. IEEE Transactions on Image Processing 31:6548--6561

\bibitem[{Wu et~al(2017)Wu, Zheng, Yu, Gong, and Lai}]{SYSU-MM01}
Wu A, Zheng WS, Yu HX, Gong S, Lai J (2017) Rgb-infrared cross-modality person re-identification. In: Proceedings of the IEEE international conference on computer vision, pp 5380--5389

\bibitem[{Wu et~al(2023)Wu, Liu, Su, Shi, and Tang}]{CAL}
Wu J, Liu H, Su Y, Shi W, Tang H (2023) Learning concordant attention via target-aware alignment for visible-infrared person re-identification. In: Proceedings of the IEEE/CVF International Conference on Computer Vision, pp 11122--11131

\bibitem[{Wu et~al(2021)Wu, Dai, Chen, Lin, Wu, Huang, Zhong, and Ji}]{MPANet}
Wu Q, Dai P, Chen J, Lin CW, Wu Y, Huang F, Zhong B, Ji R (2021) Discover cross-modality nuances for visible-infrared person re-identification. In: Proceedings of the IEEE/CVF Conference on Computer Vision and Pattern Recognition (CVPR), pp 4330--4339

\bibitem[{Wu and Ye(2023)}]{PGM}
Wu Z, Ye M (2023) Unsupervised visible-infrared person re-identification via progressive graph matching and alternate learning. In: Proceedings of the IEEE/CVF Conference on Computer Vision and Pattern Recognition, pp 9548--9558

\bibitem[{Yang et~al(2022)Yang, Ye, Chen, and Wu}]{ADCA}
Yang B, Ye M, Chen J, Wu Z (2022) Augmented dual-contrastive aggregation learning for unsupervised visible-infrared person re-identification. In: Proceedings of the 30th ACM International Conference on Multimedia, Association for Computing Machinery, New York, NY, USA, MM '22, p 2843–2851, \doi{10.1145/3503161.3548198}, \urlprefix\url{https://doi.org/10.1145/3503161.3548198}

\bibitem[{Yang et~al(2023{\natexlab{a}})Yang, Chen, Chen, and Ye}]{DCCL}
Yang B, Chen J, Chen C, Ye M (2023{\natexlab{a}}) Dual consistency-constrained learning for unsupervised visible-infrared person re-identification. IEEE Transactions on Information Forensics and Security

\bibitem[{Yang et~al(2023{\natexlab{b}})Yang, Chen, Ma, and Ye}]{TAA}
Yang B, Chen J, Ma X, Ye M (2023{\natexlab{b}}) Translation, association and augmentation: Learning cross-modality re-identification from single-modality annotation. IEEE Transactions on Image Processing

\bibitem[{Yang et~al(2023{\natexlab{c}})Yang, Chen, and Ye}]{GUR}
Yang B, Chen J, Ye M (2023{\natexlab{c}}) Towards grand unified representation learning for unsupervised visible-infrared person re-identification. In: Proceedings of the IEEE/CVF International Conference on Computer Vision (ICCV), pp 11069--11079

\bibitem[{Yang et~al(2024)Yang, Chen, and Ye}]{sdcl}
Yang B, Chen J, Ye M (2024) Shallow-deep collaborative learning for unsupervised visible-infrared person re-identification. In: Proceedings of the IEEE/CVF Conference on Computer Vision and Pattern Recognition, pp 16870--16879

\bibitem[{Yang et~al(2025)Yang, Hu, and Hu}]{PCAL-TIFS25}
Yang Y, Hu W, Hu H (2025) Progressive cross-modal association learning for unsupervised visible-infrared person re-identification. IEEE Transactions on Information Forensics and Security

\bibitem[{Ye et~al(2020)Ye, Shen, J.~Crandall, Shao, and Luo}]{DDAG}
Ye M, Shen J, J~Crandall D, Shao L, Luo J (2020) Dynamic dual-attentive aggregation learning for visible-infrared person re-identification. In: Computer Vision--ECCV 2020: 16th European Conference, Glasgow, UK, August 23--28, 2020, Proceedings, Part XVII 16, Springer, pp 229--247

\bibitem[{Ye et~al(2021{\natexlab{a}})Ye, Ruan, Du, and Shou}]{CA}
Ye M, Ruan W, Du B, Shou MZ (2021{\natexlab{a}}) Channel augmented joint learning for visible-infrared recognition. In: Proceedings of the IEEE/CVF International Conference on Computer Vision, pp 13567--13576

\bibitem[{Ye et~al(2021{\natexlab{b}})Ye, Shen, Lin, Xiang, Shao, and Hoi}]{agw}
Ye M, Shen J, Lin G, Xiang T, Shao L, Hoi SCH (2021{\natexlab{b}}) Deep learning for person re-identification: A survey and outlook. \eprint{2001.04193}

\bibitem[{Zhang et~al(2022{\natexlab{a}})Zhang, Lai, Liu, Huang, and Han}]{zhang2022fmcnet}
Zhang Q, Lai C, Liu J, Huang N, Han J (2022{\natexlab{a}}) Fmcnet: Feature-level modality compensation for visible-infrared person re-identification. In: Proceedings of the IEEE/CVF Conference on Computer Vision and Pattern Recognition, pp 7349--7358

\bibitem[{Zhang et~al(2021)Zhang, Ge, Qiao, and Li}]{RLCC}
Zhang X, Ge Y, Qiao Y, Li H (2021) Refining pseudo labels with clustering consensus over generations for unsupervised object re-identification. In: Proceedings of the IEEE/CVF Conference on Computer Vision and Pattern Recognition, pp 3436--3445

\bibitem[{Zhang et~al(2022{\natexlab{b}})Zhang, Li, Wang, Wang, Ding, Shi, Zhang, and Wang}]{ISE}
Zhang X, Li D, Wang Z, Wang J, Ding E, Shi JQ, Zhang Z, Wang J (2022{\natexlab{b}}) Implicit sample extension for unsupervised person re-identification. In: Proceedings of the IEEE/CVF Conference on Computer Vision and Pattern Recognition, pp 7369--7378

\bibitem[{Zhang and Wang(2023)}]{DEEN}
Zhang Y, Wang H (2023) Diverse embedding expansion network and low-light cross-modality benchmark for visible-infrared person re-identification. In: Proceedings of the IEEE/CVF Conference on Computer Vision and Pattern Recognition (CVPR), pp 2153--2162

\bibitem[{Zhang et~al(2024)Zhang, Lu, Yan, Wang, and Li}]{Freq1}
Zhang Y, Lu Y, Yan Y, Wang H, Li X (2024) Frequency domain nuances mining for visible-infrared person re-identification. arXiv preprint arXiv:240102162

\bibitem[{Zhong et~al(2017)Zhong, Zheng, Cao, and Li}]{k-reciprocal}
Zhong Z, Zheng L, Cao D, Li S (2017) Re-ranking person re-identification with k-reciprocal encoding. In: Proceedings of the IEEE conference on computer vision and pattern recognition, pp 1318--1327

\bibitem[{Zou et~al(2023)Zou, Chen, Cui, Liu, and Zhang}]{DCMIP}
Zou C, Chen Z, Cui Z, Liu Y, Zhang C (2023) Discrepant and multi-instance proxies for unsupervised person re-identification. In: Proceedings of the IEEE/CVF International Conference on Computer Vision (ICCV), pp 11058--11068

\end{thebibliography}

\end{sloppypar}
\end{document}